\documentclass[10pt,twocolumn,letterpaper]{article}
\pdfoutput=1
\usepackage{iccv}
\usepackage{times}
\usepackage{epsfig}
\usepackage{graphicx}
\usepackage{amsmath}
\usepackage{amssymb}

\usepackage{algorithm}
\usepackage{algorithmic}
\newtheorem{assumption}{Assumption}
\newtheorem{proposition}{Proposition}
\usepackage{subcaption}

\usepackage[pagebackref=true,breaklinks=true,letterpaper=true,colorlinks,bookmarks=false]{hyperref}

\iccvfinalcopy 

\def\iccvPaperID{***} 

\ificcvfinal\fi

\begin{document}

\title{Continuous Indeterminate Probability Neural Network}

\author{Tao Yang \\
AI Lab, United Automotive Electronic Systems Co., Ltd. \\
Shanghai, China \\
{\tt\small tao.yang9@uaes.com} 
}

\maketitle
\ificcvfinal\thispagestyle{empty}\fi

\begin{abstract}
   This paper introduces a general model called \textbf{CIPNN} \textendash\enspace\textbf{C}ontinuous \textbf{I}ndeterminate \textbf{P}robability \textbf{N}eural \textbf{N}etwork,
   and this model is based on IPNN, which is used for discrete latent random variables. Currently, posterior of continuous latent variables
   is regarded as intractable, with the new theory proposed by IPNN this problem can be solved. 
   Our contributions are Four-fold. First, we derive the analytical solution of the posterior calculation of continuous latent random variables and propose a general classification model (CIPNN).
   Second, we propose a general auto-encoder called CIPAE \textendash\enspace Continuous Indeterminate Probability Auto-Encoder, the decoder part is not a neural network and uses a fully probabilistic inference model for the first time. 
   Third, we propose a new method to visualize the latent random variables, we use one of N dimensional latent variables as a decoder to reconstruct the input image, which can work even for classification tasks,
   in this way, we can see what each latent variable has learned.
   Fourth, IPNN has shown great classification capability, CIPNN has pushed this classification capability to infinity.
   Theoretical advantages are reflected in experimental results. (Source code: https://github.com/Starfruit007/cipnn)
\end{abstract}


Although recent breakthroughs demonstrate that neural networks are remarkably adept at
natural language processing~\cite{cipnn:Transformer,cipnn:bert,cipnn:instructgpt}, image processing~\cite{cipnn:resnet}, neural networks are still black-box for human~\cite{cipnn:blackbox_NN}, cognitive scientists and neuroscientist have argued that neural
networks are limited in their ability to represent variables and data structures~\cite{cipnn:NN_limits_1,cipnn:NN_limits_2}.
Probabilistic models are mathematical descriptions of various natural and artificial phenomena learned from data,
they are useful for understanding such phenomena, for prediction of unknowns in the future, and for various forms of assisted or automated decision making~\cite{cipnn:VAE_2019}.

Deep Latent Variable Models (DLVMs) is a probabilistic model and can refer to the use of neural networks to perform
latent variable inference~\cite{cipnn:DLVMs_Tutorial}. Currently, the posterior calculation is regarded as intractable~\cite{cipnn:VAE,cipnn:VAE_2019},
and the variational inference method is used for efficient approximate posterior inference~\cite{cipnn:VAE,cipnn:VAE2,cipnn:VAE3}.

IPNN \textendash\enspace Indeterminate Probability Neural Network~\cite{cipnn:ipnn} proposed a new theory, which is used to derive the analytical solution of the posterior calculation of discrete random variables.
However, IPNN need predefine the sample space of each discrete random variable (called `split shape' in IPNN), it is sometimes hard to define a proper sample space for an unknown dataset.
For CIPNN, the sample space of each continuous random variable is infinite, this issue will not exit in CIPNN\@.

The rest of this paper is organized as follows: 
In Sec.~\ref{sec:related_work}, related work of VAE and IPNN is introduced.
In Sec.~\ref{sec:cipnn}, the analytical solution of CIPNN is derived and the regularization method is discussed.
In Sec.~\ref{sec:cipae}, CIPAE is derived and we propose a new method to visualize each latent variable. 
In Sec.~\ref{sec:training}, we discuss the training strategy, and two common training setups are discussed: CIPNN and CIPAE are combined together for better evaluation of classification and auto-encoder tasks. 
In Sec.~\ref{sec:experiment}, CIPNN and CIPAE are evaluated and the latent variables are visualized with our new proposed method.
Finally, we put forward some future research ideas and conclude the paper in  Sec.~\ref{sec:conclusion}.

\section{Related Work}\label{sec:related_work}

\subsection{VAE}

Modern machine learning and statistical applications require large scale inference in complex models, the inference model are
regarded as intractable and either Markov Chain Monte Carlo (MCMC)~\cite{cipnn:monte_carlo} 
or variational Bayesian inference~\cite{cipnn:variational_method} are used as approximate solutions~\cite{cipnn:VAE2}.
VAE~\cite{cipnn:VAE} proposes an estimator of the variational lower bound for efficient approximate inference with continuous latent variables. 
DARN method is generative auto-encoder capable of learning hierarchies of distributed representations from data, and their method applies to binary latent variables~\cite{cipnn:VAE_binary}.
In concurrent work of VAE, two later independent papers proposed equivalent algorithms~\cite{cipnn:VAE2,cipnn:VAE3}, which provides an additional perspective on VAE and the latter work applies also the same reparameterization method.
Two methods proposed by VAE are also used to realize our analytical solution: the reparameterization trick for making the model differentiable and the KL divergence term for regularization.

VAEs have been used for many tasks such as image generation~\cite{cipnn:VAE_use_image}, anomaly detection~\cite{cipnn:VAE_use_anomaly} and de-noising tasks~\cite{cipnn:VAE_use_denoise}~\cite{cipnn:VAE_use_summary}.
The drawback of auto-encoder is its strong tendency to over-fit~\cite{cipnn:overfit_AE}, as it is solely trained to encode and
decode with as little loss as possible regardless of how the latent space is organized~\cite{cipnn:overfit_VAE}, VAE has been developed as 
an effective solutions~\cite{cipnn:overfit_AE,cipnn:EEG_2}, e.g. VAEs has been used in EEG classification tasks to learn robust features~\cite{cipnn:EEG_0,cipnn:EEG_1,cipnn:EEG_2,cipnn:EEG_3}.

The framework of our CIPAE is almost the same as that of VAE,
the only difference is that VAE uses neural network as the approximate solution of decoder, 
while CIPAE uses probabilistic model as the analytical solution of decoder. 

\subsection{IPNN}

Let $X \in \{x_{1},x_{2},\dots,x_{n} \}$ be training samples ($x_{k}$ is understood as ID of random experiment \textendash\enspace select one train sample) and $Y \in \{y_{1},y_{2},\dots,y_{m} \}$ consists of m discrete labels (or classes),
$P(y_{l}|x_{k})=y_{l}(k)$ describes the label of sample $x_{k}$.
For prediction, the posterior of the label for a given new input sample $x_{t}$ 
is formulated as $P^{Z} \left ( y_{l} \mid x_{t} \right )$, 
superscript $Z$ stands for the medium \textendash\enspace model outputted N-dimensional random variables $Z = \left ( z^{1},z^{2},\dots,z^{N} \right )$, 
via which we can infer label $y_{l}, l=1,2,\dots,m$.

The analytical solution of the posterior is as bellow~\cite{cipnn:ipnn}:

\begin{equation}
    \label{eq:P_Y_X_via_Z}
    P^{Z} \left ( y_{l} \mid x_{t} \right ) 
    = \int\limits_{Z}  \left (P \left (y_{l} \mid Z \right )
    \cdot \prod_{i=1}^{N} P\left ( z^{i}\mid x_{t}   \right ) \right )
    \end{equation}

Where,

\begin{equation}
    \label{eq:P_Y_Z}
    P \left (y_{l} \mid Z \right ) 
    = \frac{ {\textstyle \sum_{k=1}^{n}} \left (  P\left (y_{l}\mid x_{k}\right )\cdot {\textstyle \prod_{i=1}^{N}}P\left ( z^{i}\mid x_{k}   \right ) \right ) }
    {{\textstyle \sum_{k=1}^{n}} {\textstyle \prod_{i=1}^{N}} P\left ( z^{i}\mid x_{k}   \right )}
    \end{equation}

\section{CIPNN}\label{sec:cipnn}
\subsection{Continuous Indeterminate Probability}\label{sec:cip}

\begin{figure}[htbp]
    \centerline{\includegraphics[width=1\linewidth]{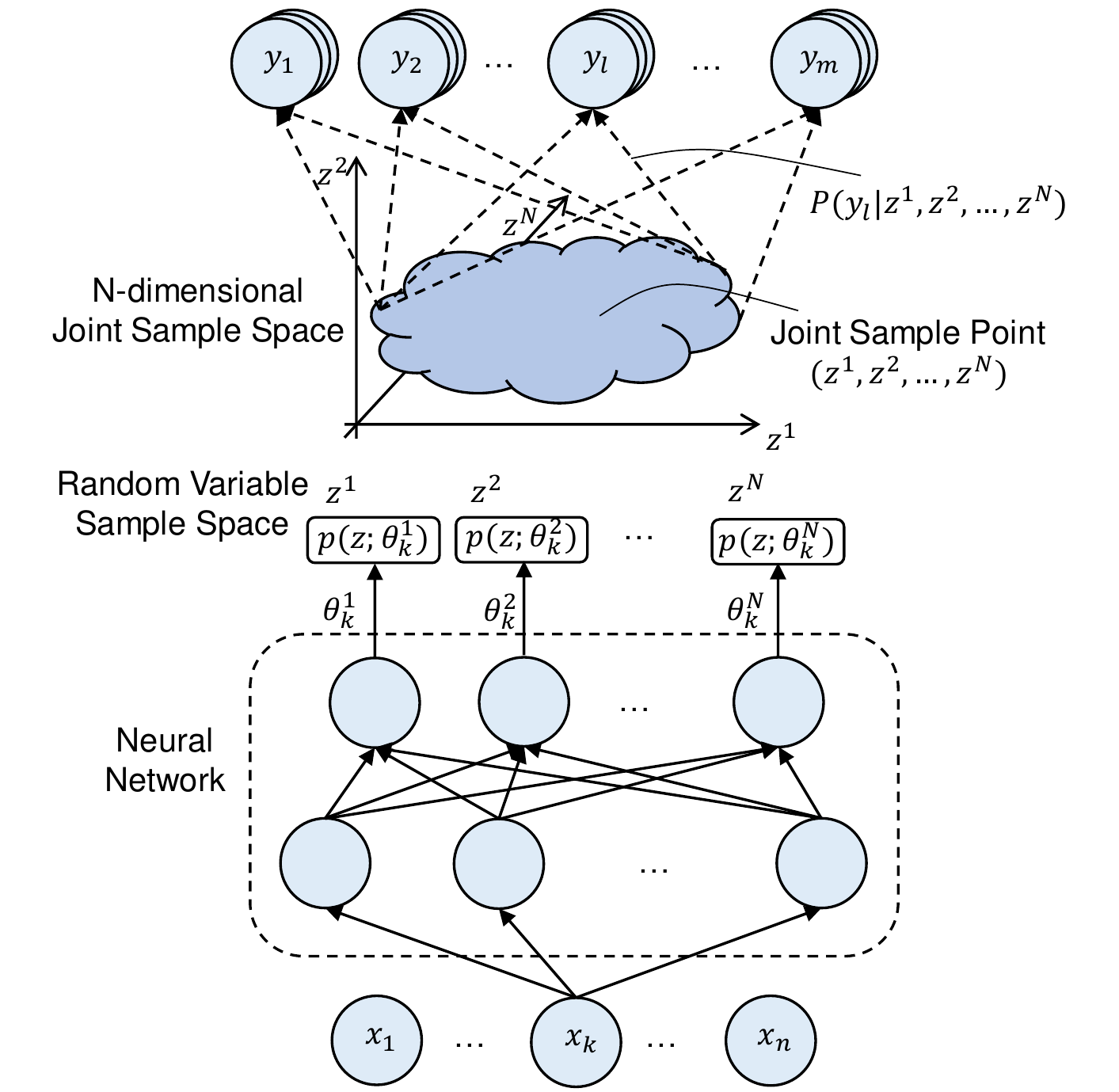}}
    \caption{CIPNN \textendash\enspace model architecture. Where
    $P \left (y_{l} \mid z^{1},z^{2},\dots,z^{N} \right )$ is statistically calculated, not model weights.}
    \label{fig:ModelArchitecture}
    \end{figure}

\figurename{~\ref{fig:ModelArchitecture}} shows CIPNN model architecture, the neural network is used to output the parameter $\theta$ of
some prior distribution of continuous random variable $z^{i}, i = 1,2,\dots,N$. 
All the random variables together form the N-dimensional joint sample space, marked as $Z = \left ( z^{1},z^{2},\dots,z^{N} \right )$, and the whole joint sample space are fully connected 
with all labels $Y \in \{y_{1},y_{2},\dots,y_{m} \}$ via conditional probability $P\left (y_{l} \mid z^{1},z^{2},\dots,z^{N} \right )$.

For each continuous random variable $z^{i}$, the indeterminate probability is formulated as: 

\begin{equation}
    \label{eq:P_z}
    P\left ( z^{i} \mid x_{k} \right ) 
    = p \left ( z;\theta_{k}^{i}   \right ), i = 1,2,\dots,N.
    \end{equation}

Where $z$ is generated from some prior distribution with parameter $\theta_{k}^{i}$,
and $p \left ( z;\theta_{k}^{i}   \right )$ is also the density function of $P\left ( z^{i} \mid x_{k} \right )$.

Substitute $P(y_{l}|x_{k})=y_{l}(k)$ and  Eq.~\eqref{eq:P_z} into Eq.~\eqref{eq:P_Y_X_via_Z}:

\begin{multline}
    \label{eq:P_Y_X_via_Z_1}
    P^{Z} \left ( y_{l} \mid x_{t} \right ) \\
    \begin{aligned}
    &=\int\limits_{Z}  \left ( \frac{ \sum_{k=1}^{n} \left (  y_{l}(k) \prod_{i}^{N}p \left ( z;\theta_{k}^{i}   \right ) \right ) }
    { \sum_{k=1}^{n} \left (  \prod_{i}^{N}p \left ( z;\theta_{k}^{i}   \right ) \right )} 
    \prod_{i}^{N}p\left (z; \theta_{t}^{i}   \right ) \right ) \\
    &=\mathbb{E}_{z^\sim p \left ( z;\theta_{t}^{i}   \right )}  \left ( \frac{ \sum_{k=1}^{n} \left (  y_{l}(k)\cdot \prod_{i}^{N}p \left ( z;\theta_{k}^{i}   \right ) \right ) }
    { \sum_{k=1}^{n} \left (  \prod_{i}^{N}p \left ( z;\theta_{k}^{i}   \right ) \right )} \right )
    \end{aligned}
    \end{multline}

As the integration over $Z$ is complicated, $P^{Z} \left ( y_{l} \mid x_{t} \right )$ is rewritten as expectation, we can then
use Monte Carlo method~\cite{cipnn:monte_carlo} to make an approximate estimation. However, a directly sampling $z$ from distribution $p \left ( z;\theta_{t}^{i}   \right )$
will make the exception not differentiable. Hence, we use the reparameterization trick~\cite{cipnn:VAE}: let $\varepsilon\sim p \left ( \varepsilon \right )$ be some random noise,
and define a mapping function $z = g(\varepsilon,\theta)$, so $p \left ( z;\theta_{k}^{i}   \right )$ can be rewritten as $p \left ( g(\varepsilon,\theta);\theta_{k}^{i}   \right )$.

Therefore, together with Monte Carlo method, the above function can be further formulated as:

\begin{multline}
    \label{eq:P_Y_X_via_Z_2}
    P^{Z} \left ( y_{l} \mid x_{t} \right ) \\
    \begin{aligned}
    &=\mathbb{E}_{\varepsilon\sim p \left ( \varepsilon \right )}  \left ( \frac{ \sum_{k=1}^{n} \left (  y_{l}(k) \prod_{i}^{N}p \left ( g(\varepsilon,\theta_{t}^{i});\theta_{k}^{i}   \right ) \right ) }
    { \sum_{k=1}^{n} \left (  \prod_{i}^{N}p \left ( g(\varepsilon,\theta_{t}^{i});\theta_{k}^{i}   \right ) \right )} \right ) \\
    &\approx \frac{1}{C}\sum_{c=1}^{C} \left ( \frac{ \sum_{k=1}^{n} \left (  y_{l}(k)\cdot \prod_{i}^{N}p \left ( g(\varepsilon_{c},\theta_{t}^{i});\theta_{k}^{i}   \right ) \right ) }
    { \sum_{k=1}^{n} \left (  \prod_{i}^{N}p \left ( g(\varepsilon_{c},\theta_{t}^{i});\theta_{k}^{i}   \right ) \right )} \right )
    \end{aligned}
    \end{multline}
    
Where $\varepsilon_{c}\sim p \left ( \varepsilon \right )$.

Take, for example, the univariate Gaussian case: let $P\left ( z^{i} \mid x_{k} \right ) = \mathcal{N} \left ( z;\mu_{k}^{i} ,\sigma_{k}^{i}  \right )$,
and $\varepsilon_{c}\sim \mathcal{N} \left (0,1 \right )$, the reparameterization mapping function 
is $z = \mu+\sigma\varepsilon $, the above function can be rewritten as:

\begin{multline}
    \label{eq:P_Y_X_via_Z_norm}
    P^{Z} \left ( y_{l} \mid x_{t} \right ) \\
    \begin{aligned}
    &\approx \frac{1}{C}\sum_{c=1}^{C} \left ( \frac{ \sum_{k=1}^{n} \left (  y_{l}(k) \cdot\prod_{i}^{N} \mathcal{N} \left ( g\left ( \varepsilon_{c},\theta_{t}^{i}  \right );\theta_{k}^{i}  \right )  \right ) }
    {\sum_{k=1}^{n} \left ( \prod_{i}^{N} \mathcal{N} \left ( g\left ( \varepsilon_{c},\theta_{t}^{i} \right );\theta_{k}^{i}  \right ) \right ) } \right )
    \end{aligned}
    \end{multline}
    
Where $\theta_{k}^{i}:=(\mu_{k}^{i} ,\sigma_{k}^{i})$, $g\left ( \varepsilon_{c},\theta_{t}^{i} \right )=\mu_{t}^{i}+\sigma_{t}^{i}\cdot\varepsilon_{c} $ 
and $\varepsilon_{c}\sim \mathcal{N} \left (0,1 \right )$.

We use cross entropy as main loss function:
\begin{equation}
    \label{eq:main_loss}
    \mathcal{L}_{1} =  -{\textstyle \sum_{l=1}^{m}} \left (  y_{l}(t)\cdot  \log P^{Z} \left ( y_{l} \mid x_{t} \right )\right )
\end{equation}

\subsection{Regularization}

The sufficient and necessary condition of achieving global minimum is already proved in IPNN~\cite{cipnn:ipnn}, which is also valid for continuous latent variables:

\begin{proposition}
    \label{pps:convergence}
    For $P(y_{l}|x_{k})=y_{l}(k) \in \{0,1\}$ hard label case, CIPNN converges to global minimum only when $P\left (y_{l}| z^{1},z^{2},\dots,z^{N} \right ) \to 1,
    \text{ for } \prod_{i}^{N}p \left ( z;\theta_{k}^{i}   \right )  > 0$. 
    
    In other word, each N-dimensional joint sample area (collection of adjacent joint sample points) corresponds to an unique category.
    However, a category can correspond to one or more joint sample areas. 
\end{proposition}

According to above proposition, the reduction of training loss will minimize the overlap between conditional joint distribution $\prod_{i}^{N}p \left ( z;\theta_{k}^{i}   \right )$ of each category. 
For Gaussian distribution, the variance will be close to zero, 
and the conditional joint distribution of each category will be far away from each other.
This will cause over-fitting problem~\cite{cipnn:overfit_VAE,cipnn:overfit_AE}, we have follow up assumption to avoid over-fitting problem:

\begin{assumption}
    \label{asm:overfitting}
    For Gaussian distribution, if the distance between the centers of any two adjacent categories' conditional joint distribution $\prod_{i}^{N}\mathcal{N}  \left ( z;\mu_{k}^{i} ,\sigma_{k}^{i}  \right )$ is equal to e.g. $6\sigma$, then the over-fitting problem can be avoided.
    
    In this way, the $6\sigma$ distance will lead to a very small overlap between each category, but the conditional joint distribution of each category will be closely connected with each other. 
    Although the N-dimensional joint sample space is infinite, the effective conditional joint distributions are in a very small space.
\end{assumption}

VAE uses an additional regularization loss to avoid the over-fitting problem~\cite{cipnn:VAE,cipnn:VAE_2019},
and there are follow up works which has proposed to strengthen this regularization term, such as $\beta$-VAE~\cite{cipnn:betaVAE,cipnn:betaVAE_understanding}, $\beta$-TCVAE~\cite{cipnn:TCVAE}, etc.
In order to realize the above assumption, we have a modification of VAE regularization loss:

\begin{multline}
    \label{eq:regular_loss}
    \mathcal{L}_{2}  = \sum_{i=1}^{N}\left (D_{KL}\left ( \mathcal{N}  \left ( z;\mu_{k}^{i} ,\sigma_{k}^{i}   \right )  \left |  \right | \mathcal{N}  \left (z;\gamma \cdot\mu_{k}^{i} ,1  \right ) \right ) \right )\\
    = \frac{1}{2} \sum_{i=1}^{N}\left ( ((1-\gamma) \cdot \mu_{k}^{i})^2 + (\sigma_{k}^{i})^2 - \log((\sigma_{k}^{i})^2)-1\right ) 
    \end{multline}

Where $N$ is the dimensionality of $Z$, regularization factor $\gamma \in [0,1]$ is a hyperparameter 
and is used to constrain the conditional joint distribution of each category to be closely connected with each other, 
impact analysis of regularization factor $\gamma$ see~\figurename{~\ref{fig:2d_z_mnist_gammas}}.

The overall loss is:
\begin{equation}
    \label{eq:overall_loss}
    \mathcal{L} =  \mathcal{L}_{1} + \mathcal{L}_{2}
\end{equation}

\section{CIPAE}\label{sec:cipae}

For image auto-encoder task, we firstly transform the pixel value to $[0,1]$ (Bernoulli distribution), and let $Y^{j} \in \{y_{1}^{j},y_{2}^{j}\}_{j=1}^{J}$, where $J$ is the number of all pixels of one image.
$P(y_{1}^{j}|x_{k})=p_{1}^{j}(k) \in [0,1]$, which describes the pixel value of image $x_{k}$ at $j^{th}$ position, and $P(y_{2}^{j}|x_{k})=p_{2}^{j}(k)=1-p_{1}^{j}(k)$.

Substitute $P(y_{l}^{j}|x_{k})$ into Eq.~\eqref{eq:P_Y_X_via_Z_norm}, we will get $P^{Z} \left ( y_{l}^{j} \mid x_{t} \right ), l = 1,2$.
In this way, the reconstructed image is formulated as:

\begin{equation}
    \label{eq:reconstructed_image}      
    \text{reconstructed image} :=\left \{ P^{Z} \left ( y_{1}^{j} \mid x_{t} \right )\right \}_{j=1}^{J}
    \end{equation}

In addition, with one (or part) of N dimensional latent variables we can also reconstruct the input image,
the reconstructed image is:\footnote{The details of applying the superscript $z^{i}$ are discussed in IPNN~\cite{cipnn:ipnn}.}

\begin{equation}
    \label{eq:reconstructed_latent_image}      
    \text{reconstructed feature} : = \left\{P^{z^{i}} \left ( y_{1}^{j} \mid x_{t} \right )\right\}_{j=1}^{J}
    \end{equation}

Where $i = 1,2,\dots,N$. In this way, we can see what each latent variable has learned.

Substitute Eq.~\eqref{eq:reconstructed_image} into Eq.~\eqref{eq:main_loss},
we can get a binary cross entropy loss:

\begin{equation}
    \label{eq:main_loss_CIPAE}      
    \mathcal{L}_{1} =  -\frac{1}{J} \sum_{j=1}^{J}\sum_{l=1}^{2}
    \left( p_{l}^{j}(t) \cdot \log P^{Z}\left ( y_{l}^{j} \mid x_{t} \right ) \right)
    \end{equation}

And substitute the above loss into Eq.~\eqref{eq:overall_loss}, we get the overall loss for auto-encoder training.

\section{Training}\label{sec:training}

In this section, we will focus on the training strategy of gaussian distribution.

\subsection{Training Strategy}

Given an input sample $x_{t}$ from a mini batch,  with a minor modification of Eq.~\eqref{eq:P_Y_X_via_Z_norm}:

\begin{equation}
    \label{eq:P_Y_X_via_A_Train}
    P^{Z} \left ( y_{l} \mid x_{t} \right ) \approx  \frac{1}{C} \sum_{c=1}^{C} \left ( \frac{ \max(H(\varepsilon_{c}) ,\epsilon ) } {\max(G(\varepsilon_{c}) ,\epsilon ) } \right )
    \end{equation}

Where stable number $\epsilon$ on the denominator is to avoid dividing zero, 
$\epsilon$ on the numerator is to have an initial value of 1. Besides,

\begin{align} 
    \label{eq:H}
    &H(\varepsilon_{c}) = \sum_{k=t_{0}}^{t_{1}} \left (  y_{l}(k) \prod_{i}^{N} \mathcal{N} \left ( \mu_{t}^{i}+\sigma_{t}^{i}\cdot\varepsilon_{c} ;\mu_{k}^{i},\sigma_{k}^{i}  \right )  \right )\\
    \label{eq:G}
    &G(\varepsilon_{c})  = \sum_{k=t_{0}}^{t_{1}} \left ( \prod_{i}^{N} \mathcal{N} \left ( \mu_{t}^{i}+\sigma_{t}^{i}\cdot\varepsilon_{c} ;\mu_{k}^{i},\sigma_{k}^{i}  \right ) \right )
\end{align}

Where $t_{0} = \max(0,t_{1}-T)$, $t_{1}$ is the number of input samples, $\varepsilon_{c}\sim \mathcal{N} \left (0,1 \right )$.
Hyperparameter T is for forgetting use, i.e., $P^{Z} \left ( y_{l} \mid x_{t} \right )$ are calculated from the recent T samples. 
The detailed algorithm implementation is shown in Algorithm~\eqref{alg:train}.

\begin{algorithm}[tb]
    \caption{CIPNN or CIPAE training}
    \label{alg:train}
    \textbf{Input}: A sample $x_{t}$ from mini-batch\\
    \textbf{Parameter}: Latent variables dimension $N$, forget number $T$, Monte Carlo number $C$, regularization factor $\gamma$, stable number $\epsilon$, learning rate $\eta$.\\
    \textbf{Output}: $P^{Z} \left ( y_{l} \mid x_{t} \right )$ 
    \begin{algorithmic}[1]
        \STATE Declare $\Theta$ as a recorder.  
        \FOR{$k=1,2,\dots$ Until Convergence}
        \STATE Use $\Theta$ to record: $y_{l},\mu_{k}^{i},\sigma_{k}^{i}, i = 1,2,\dots,N$.  
        \IF{$len(\Theta)>T$}
            \STATE Forget: Reserve recent T elements from $\Theta$
        \ENDIF
        \STATE Compute posterior with Eq.~\eqref{eq:P_Y_X_via_A_Train}: $P^{Z} \left ( y_{l} \mid x_{t} \right )$ 
        \STATE Compute loss with Eq.~\eqref{eq:overall_loss}: $\mathcal{L}(W )$
        \STATE Update model parameter: $W = W - \eta\nabla \mathcal{L}(W)$
        \ENDFOR
        \STATE \textbf{return} model and the posterior
    \end{algorithmic}
\end{algorithm}

\subsection{Training Setups}\label{sec:_train_setup}

\begin{figure}[htbp]
    \centering
        \begin{subfigure}[b]{0.49\linewidth}
                \centering
                \includegraphics[width=1\linewidth]{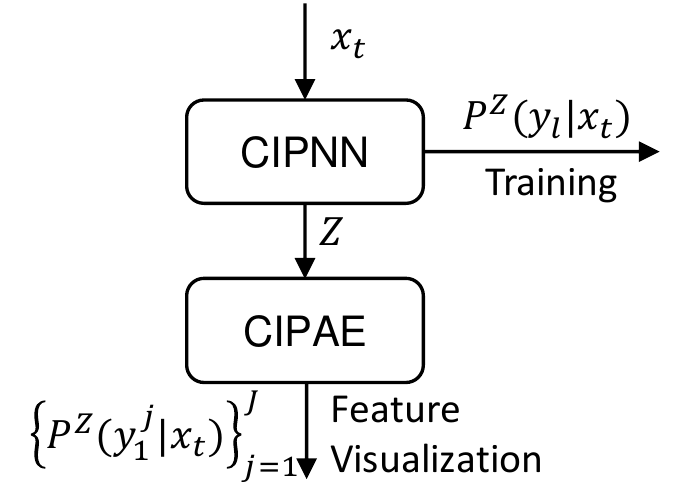}
                \caption{Classification tasks}
                \label{fig:train_process_a}
        \end{subfigure}
        \begin{subfigure}[b]{0.49\linewidth}
                \centering
                \includegraphics[width=1\linewidth]{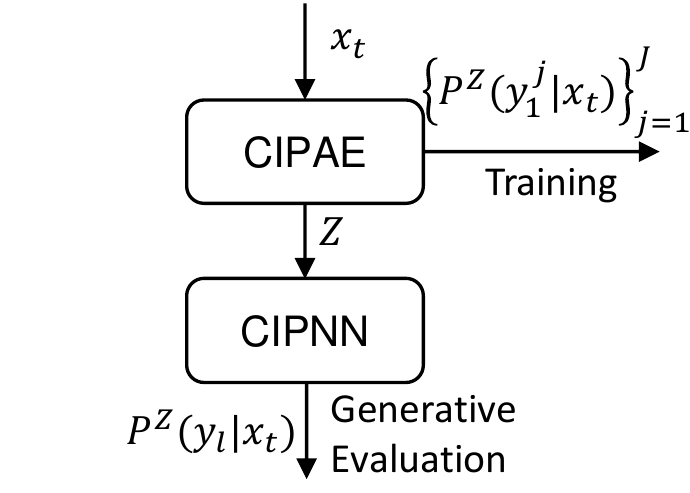}
                \caption{Auto-encoder tasks}
                \label{fig:train_process_b}
        \end{subfigure}
        \caption{Training setups for classification and auto-encoder tasks. (a) CIPNN is used to do supervised classification tasks and CIPAE is used to reconstruct
        the input image to see what each latent variable has learned. (b) CIPAE is used to do auto-encoder task and CIPNN is used to evaluate its performance.}
        \label{fig:train_process}
    \end{figure}

By comparing CIPNN and CIPAE, we can see that they can share the same neural network for a training task.
As shown in \figurename{~\ref{fig:train_process}}, the latent variables of a classification task can be visualized with CIPAE,
and we can also use CIPNN to evaluate the performance of an auto-encoder task.

\section{Experiments and Results}\label{sec:experiment}

To evaluate the effectiveness of the proposed approach, we conducted experiments on MNIST~\cite{cipnn:mnist},
Fashion-MNIST~\cite{cipnn:fashion_mnist} and Dogs-vs.-Cats-Redux~\cite{cipnn:dogs-vs-cats-redux} datasets.

Besides, VAE validated that Monte Carlo number $C$ can be set to 1 as long as the batch size is high enough (e.g. 100)~\cite{cipnn:VAE},
we will set batch size to 64, Monte Carlo number $C=2$ and forget number $T=3000$ for all the experiments in this paper.

\subsection{Results of Classification Tasks}\label{sec:classification_results}

In this section, we use train setup in~\figurename{~\ref{fig:train_process_a}} to perform different classification tasks
in order to reconstruct the latent variable to see what they have learned.

\begin{figure}[htbp]
\centering
    \centering
    \includegraphics[width=1\linewidth]{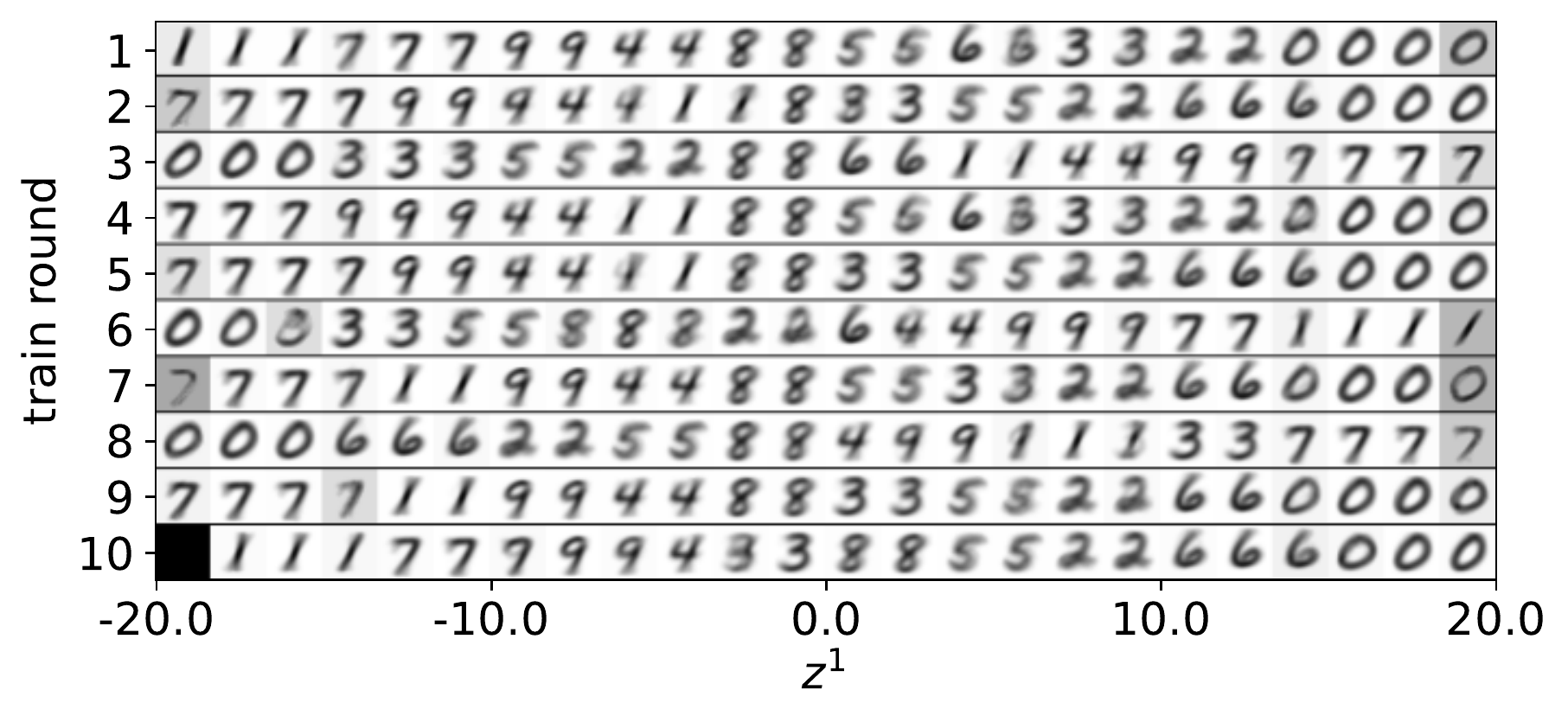}
    \caption{Reconstructed image with 1-D latent space for classification of MNIST\@: test accuracy is $93.3\pm0.5 \%, \gamma=0.95, \epsilon=1$\@.
            The training is repeated for 10 rounds with different random seeds.}
    \label{fig:1d_z_mnist}
    \end{figure}

\begin{figure*}[htbp]
    \centering
        \begin{subfigure}[c]{0.33\linewidth}
                \centering
                \includegraphics[width=1\linewidth]{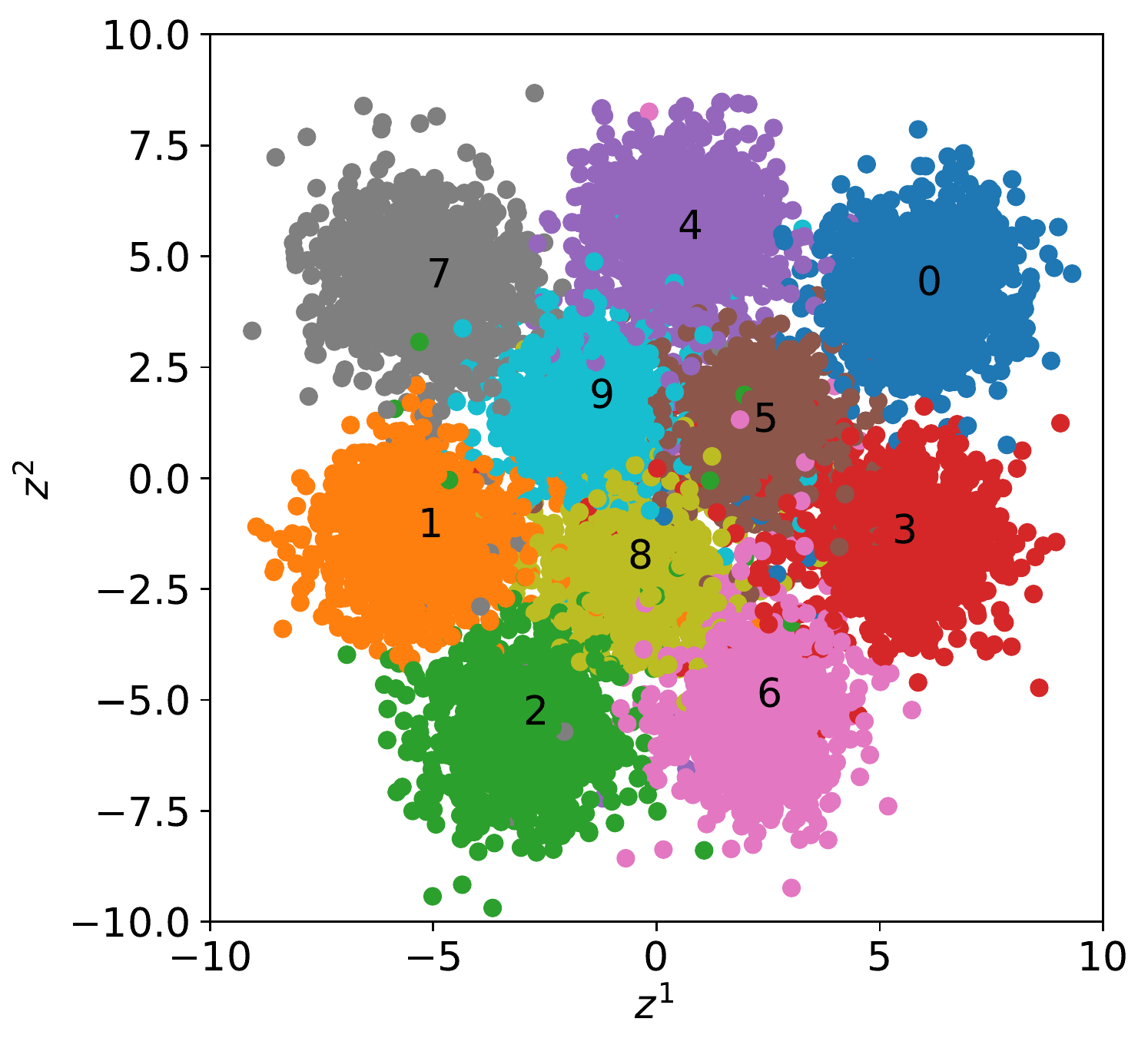}
                \caption{}
                \label{fig:2d_z_mnist_a}
        \end{subfigure}
        \begin{subfigure}[c]{0.33\linewidth}
                \centering
                \includegraphics[width=1\linewidth]{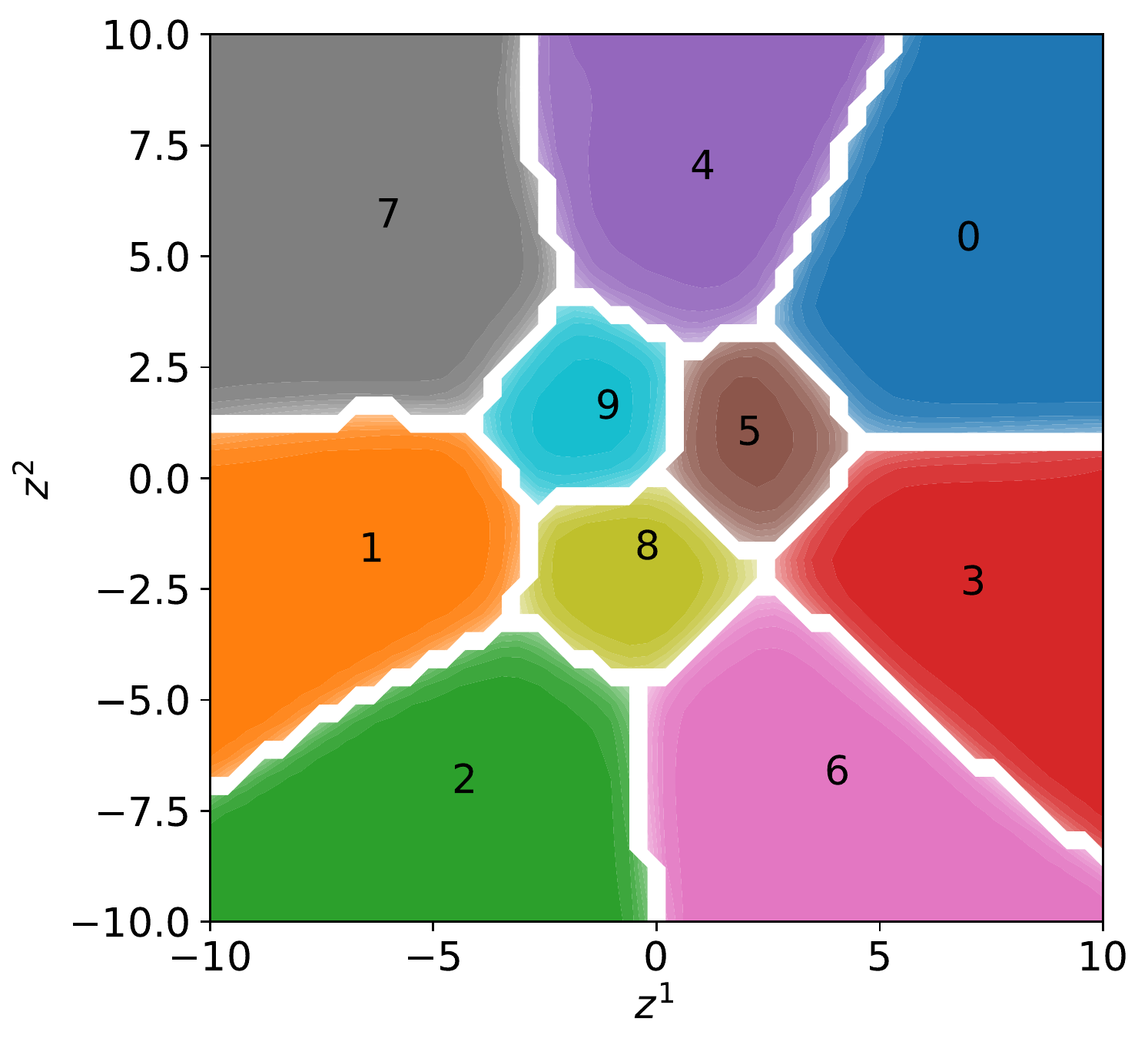}
                \caption{}
                \label{fig:2d_z_mnist_b}
        \end{subfigure}
        \begin{subfigure}[c]{0.33\linewidth}
            \centering
            \includegraphics[width=1\linewidth]{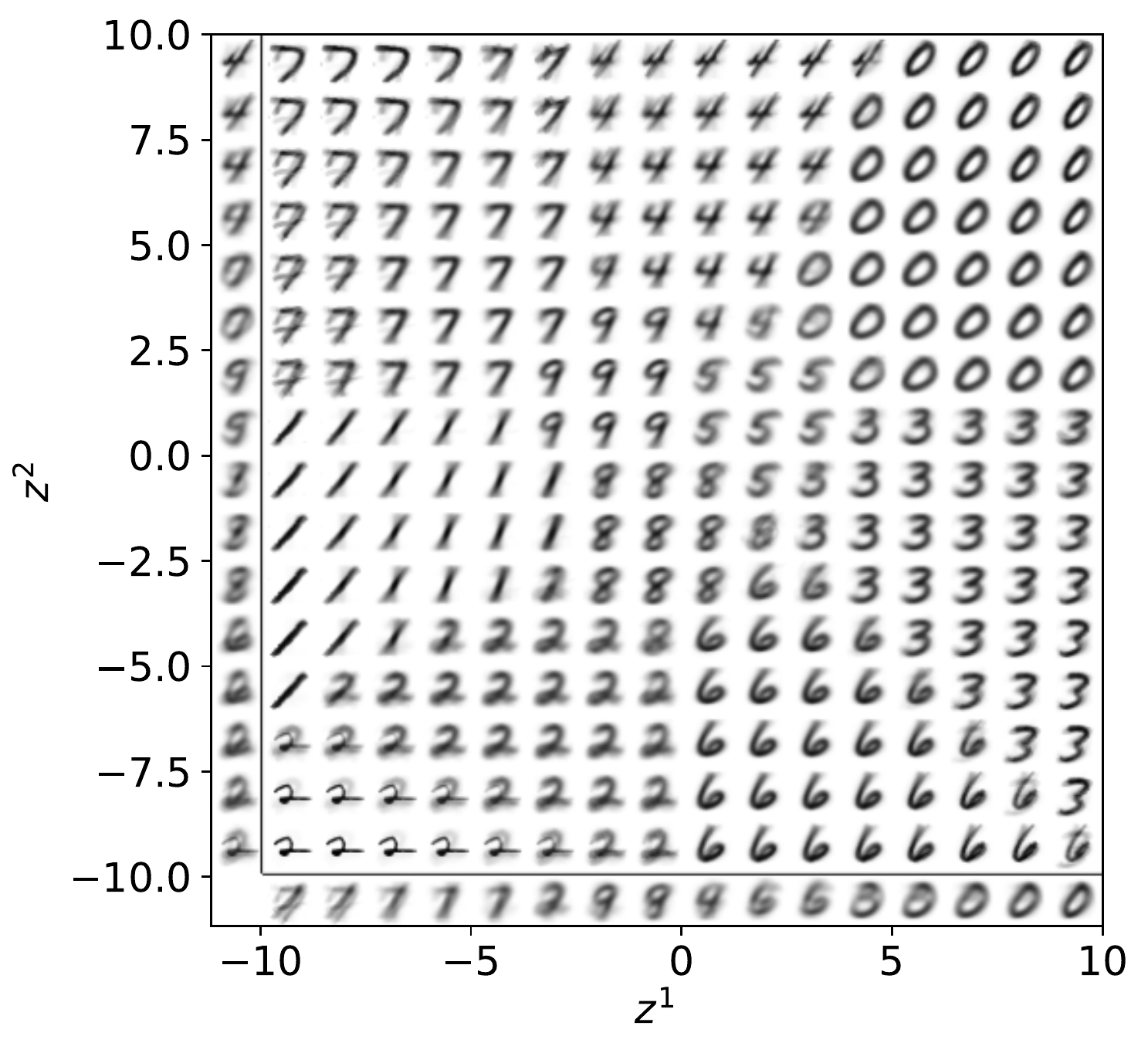}
            \caption{}
            \label{fig:2d_z_mnist_c}
        \end{subfigure}

        \begin{subfigure}[c]{0.33\linewidth}
            \centering
            \includegraphics[width=1\linewidth]{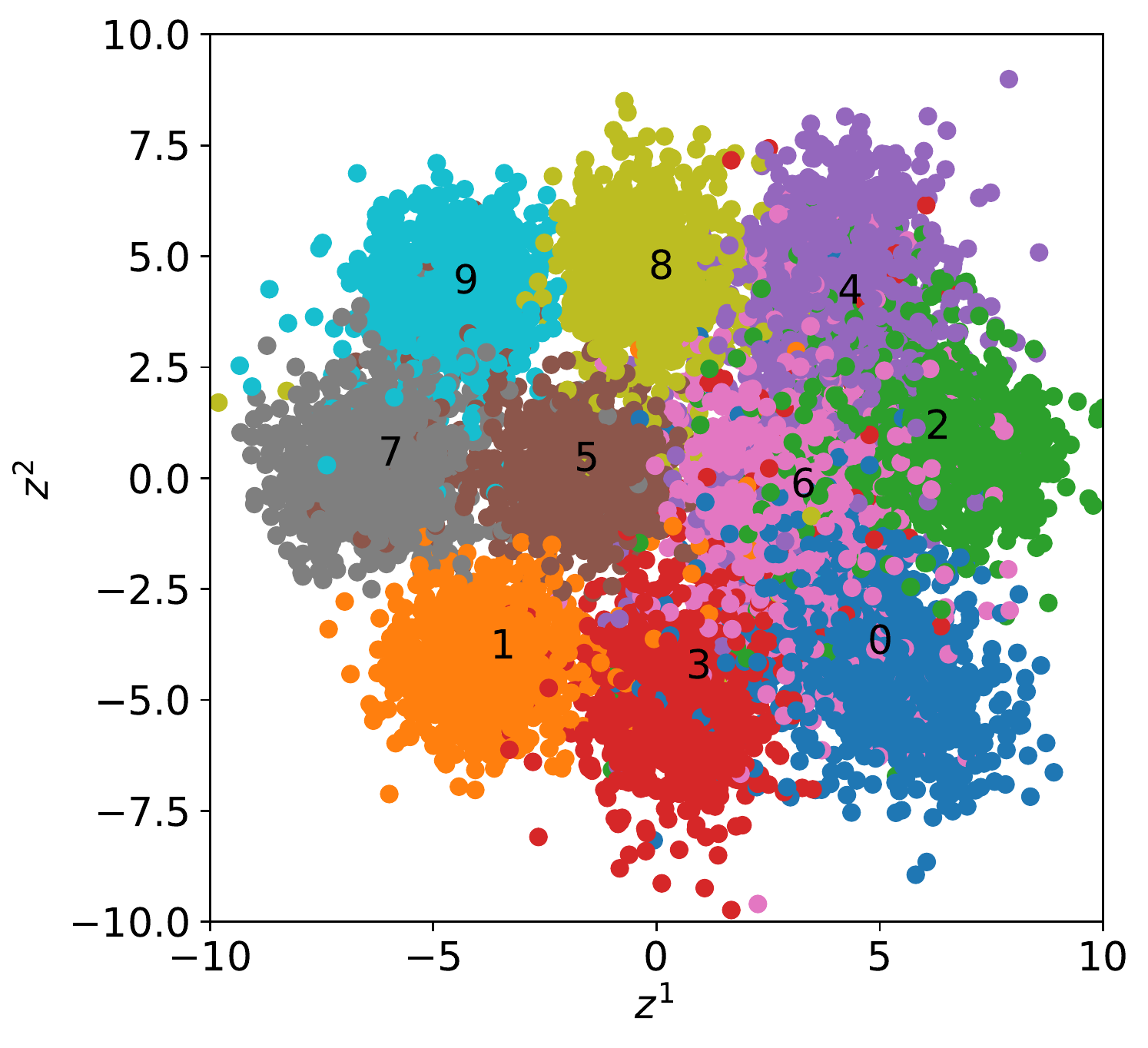}
            \caption{}
            \label{fig:2d_z_fashionmnist_a}
        \end{subfigure}
        \begin{subfigure}[c]{0.33\linewidth}
                \centering
                \includegraphics[width=1\linewidth]{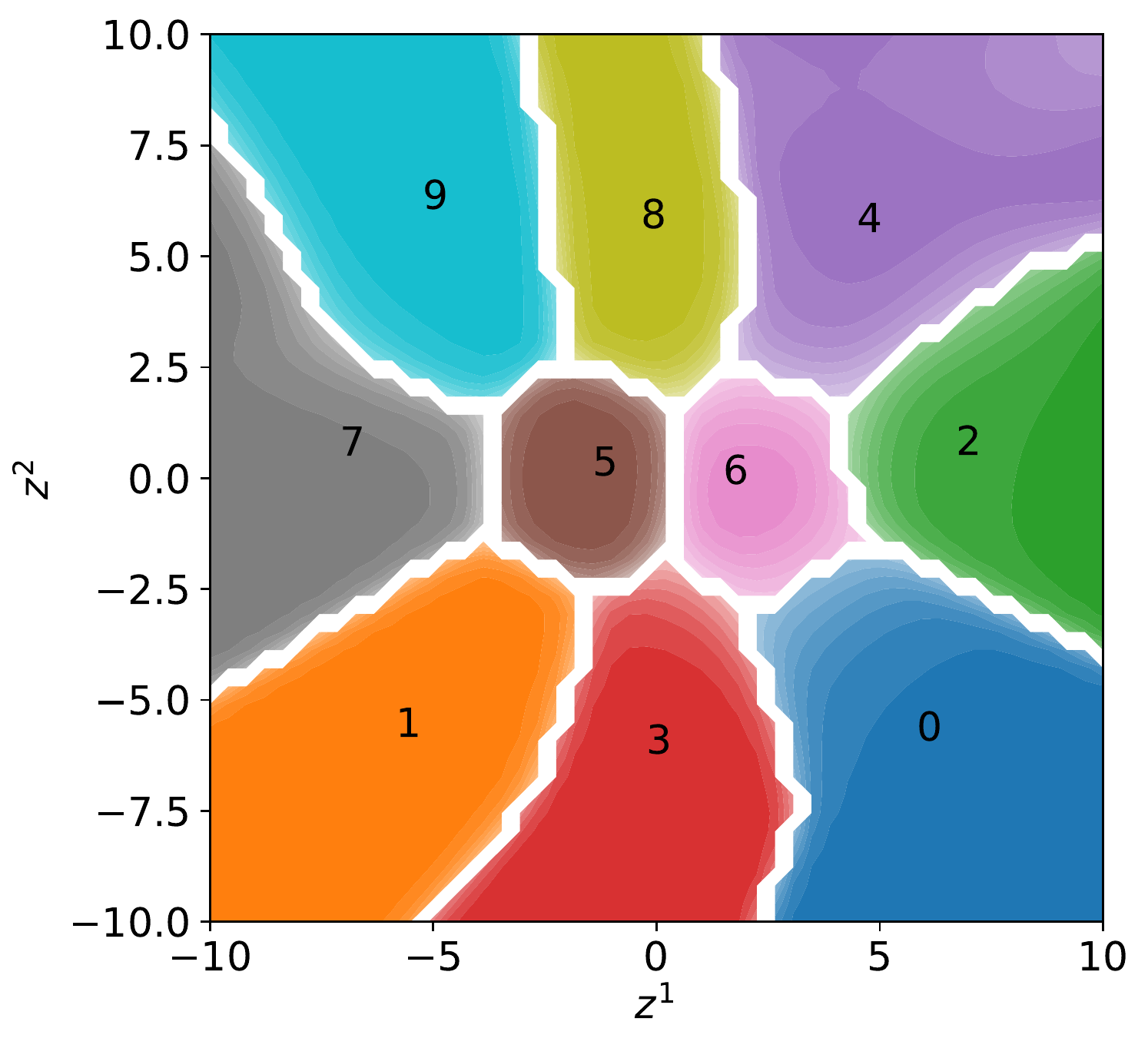}
                \caption{}
                \label{fig:2d_z_fashionmnist_b}
        \end{subfigure}
        \begin{subfigure}[c]{0.33\linewidth}
            \centering
            \includegraphics[width=1\linewidth]{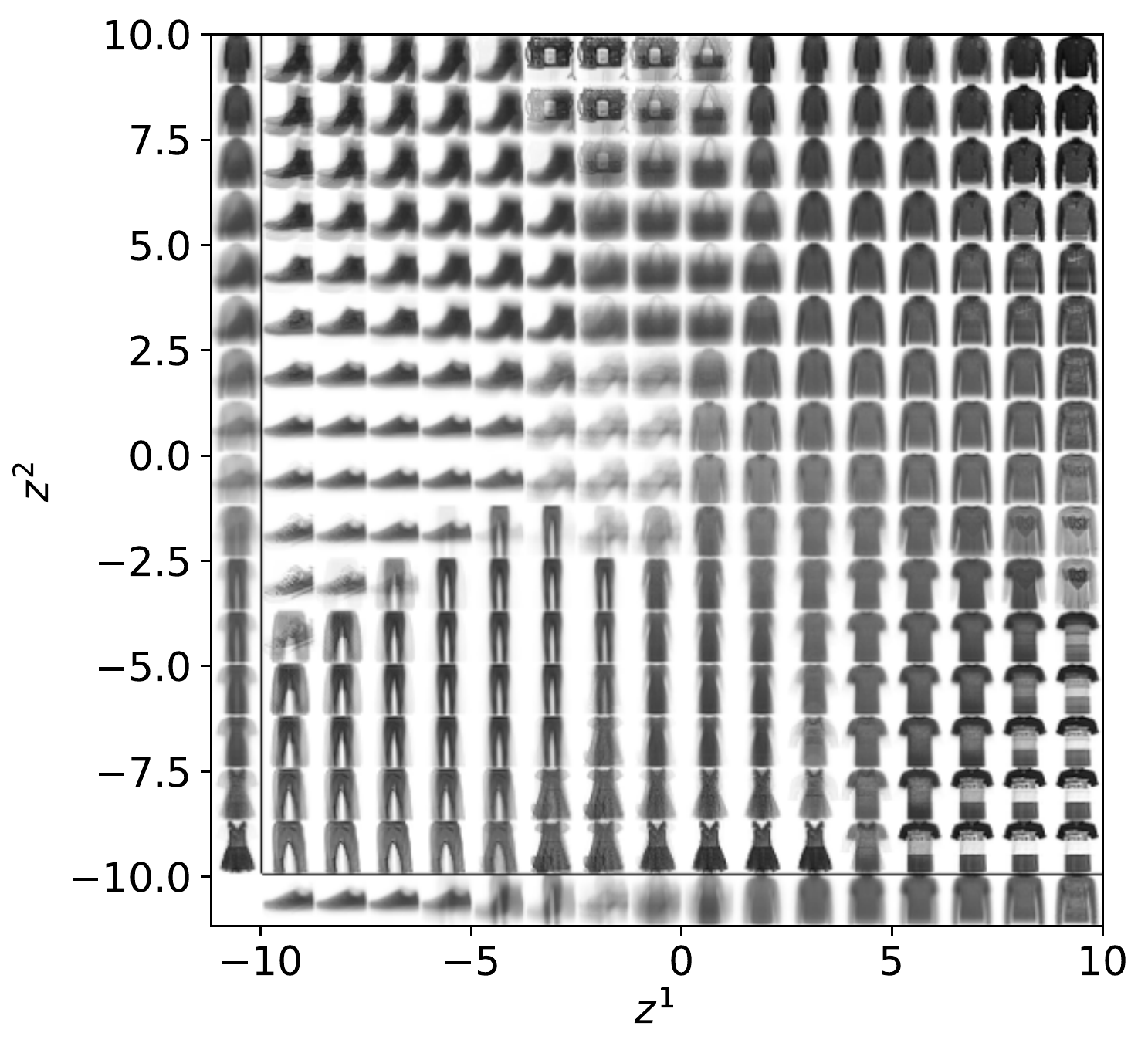}
            \caption{}
            \label{fig:2d_z_fashionmnist_c}
        \end{subfigure}

        \begin{subfigure}[c]{0.33\linewidth}
            \centering
            \includegraphics[width=1\linewidth]{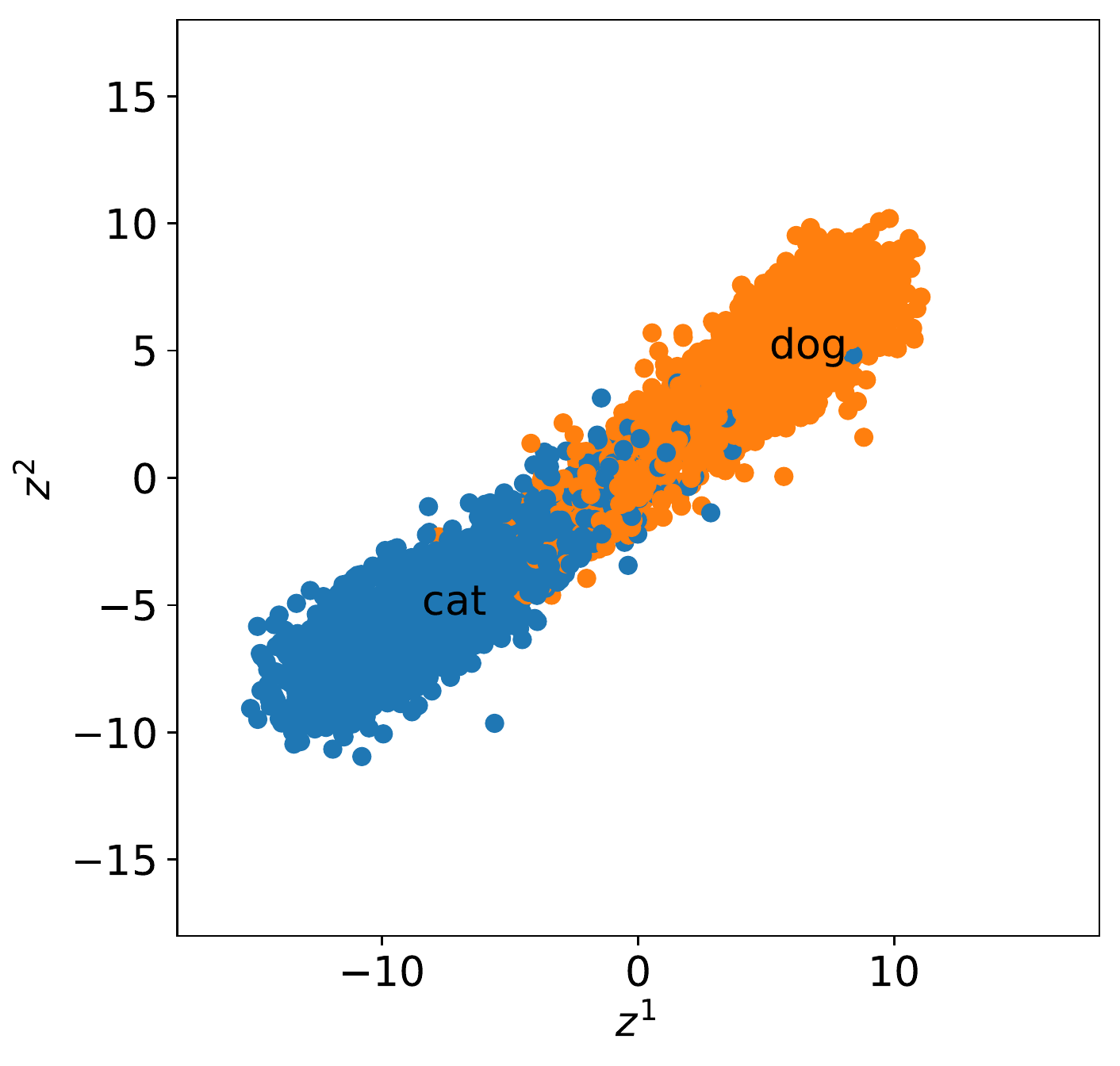}
            \caption{}
            \label{fig:2d_z_petsimages_a}
        \end{subfigure}
        \begin{subfigure}[c]{0.33\linewidth}
                \centering
                \includegraphics[width=1\linewidth]{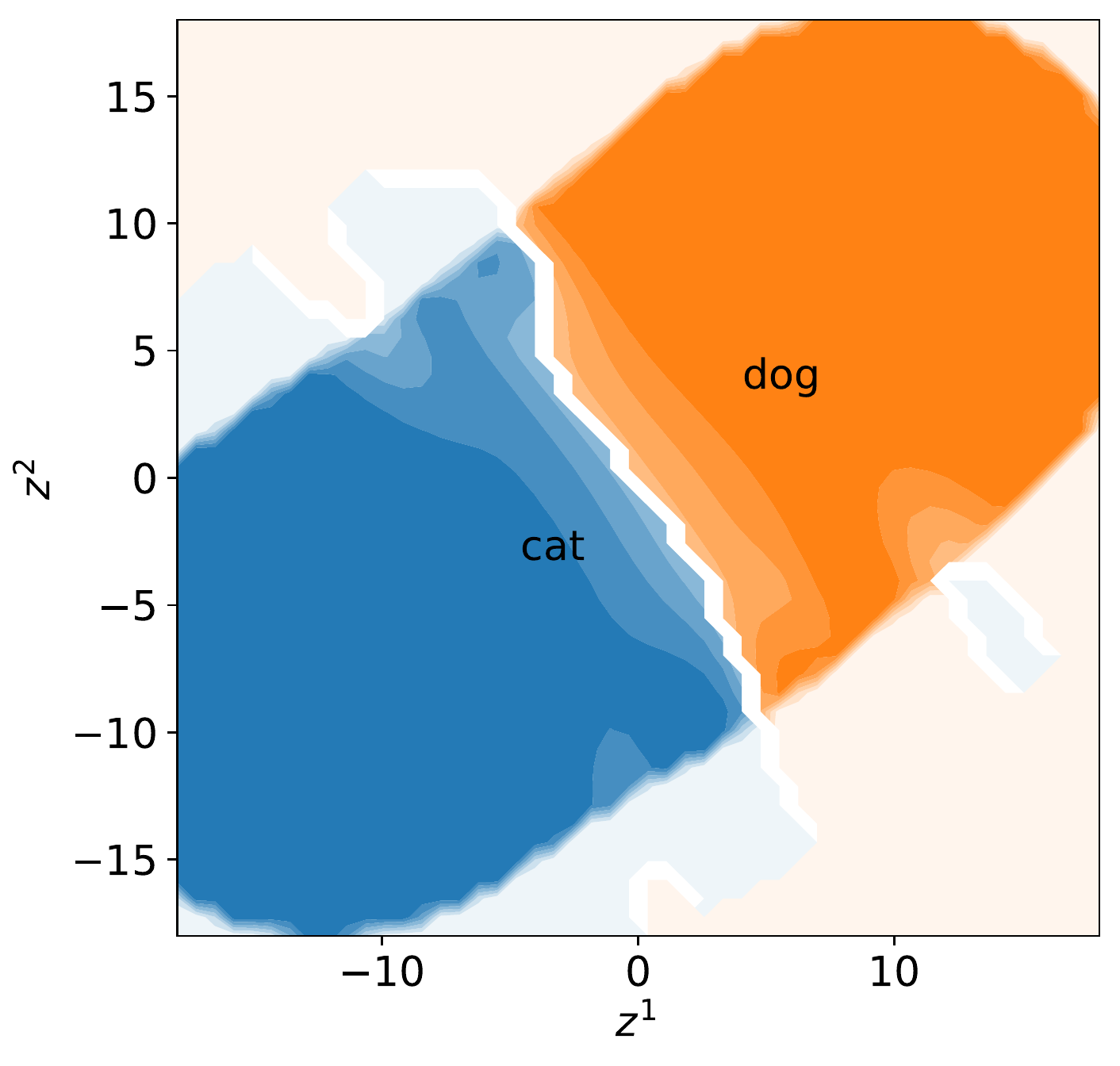}
                \caption{}
                \label{fig:2d_z_petsimages_b}
        \end{subfigure}
        \begin{subfigure}[c]{0.33\linewidth}
            \centering
            \includegraphics[width=1\linewidth]{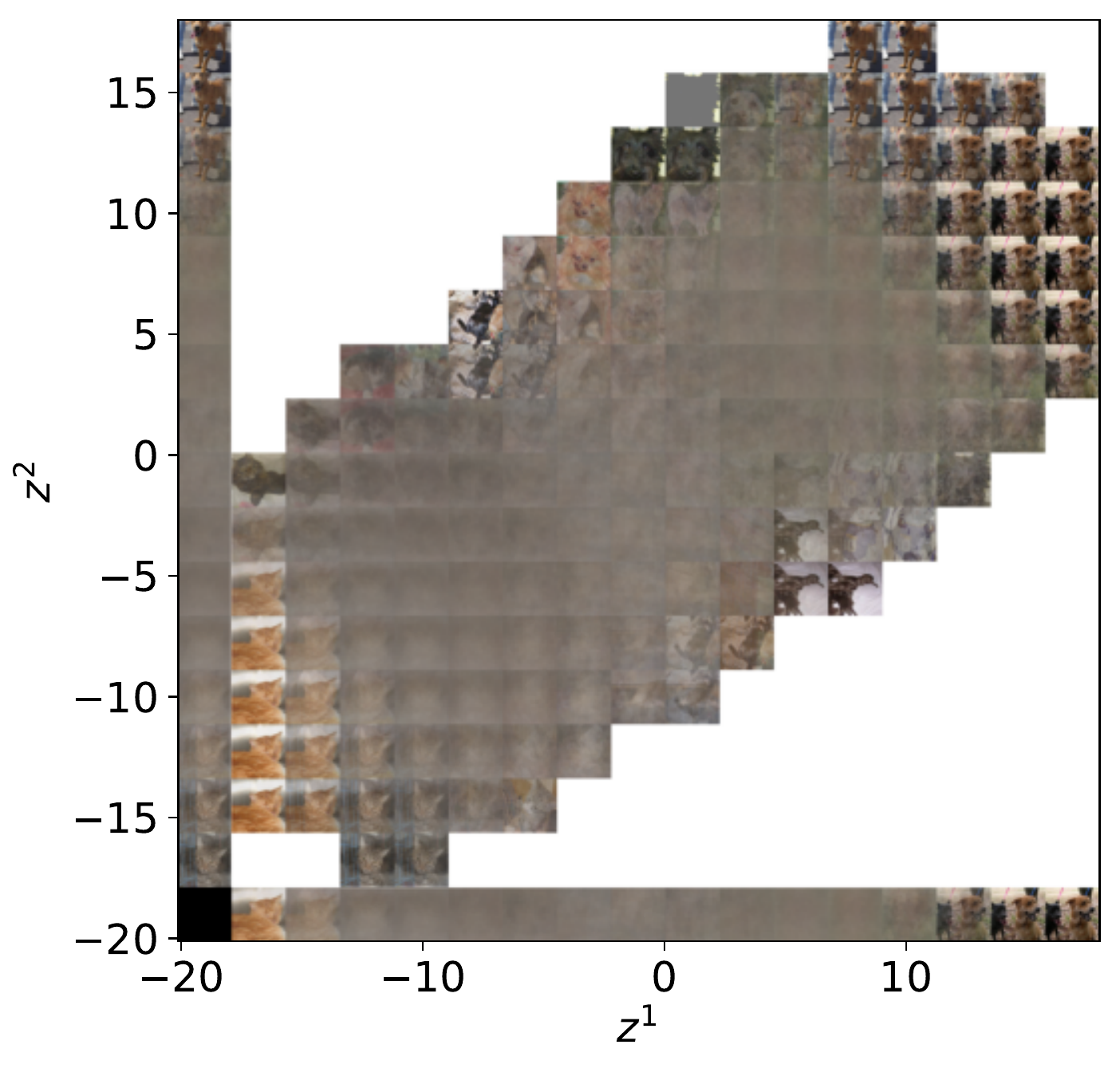}
            \caption{}
            \label{fig:2d_z_petsimages_c}
        \end{subfigure}
        \caption{Classification Results of 2-D latent space on MNIST, Fashion-MNIST and Dogs-vs.-Cats-Redux\@: 
        test accuracy is $96.1\%,87.6\%$ and $95\%$, respectively. $\gamma=0.9$ for MNIST and Fashion-MNIST,
        $\gamma=0.9999$ for Dogs-vs.-Cats-Redux, $\epsilon\approx 0$\@. 
        (a,d,g) Results of latent variables on test dataset;
        (b,e,h) Conditional probability distribution of each category $P \left (y_{l} \mid z^{1},z^{2} \right ), l = 0,2,\dots,9$. Colors represent probability value: from 1-dark to 0-light;
        (c,f,i) Reconstructed image with $(z^{1},z^{2})$, see Eq.~\eqref{eq:reconstructed_image}, 
                and image on x and y axes is reconstructed with $z^{1}$ and $z^{2}$, respectively, see Eq.~\eqref{eq:reconstructed_latent_image}.
        }
        \label{fig:2d_z_mnist}
        \end{figure*}

The results from the work~\cite{cipnn:ipnn} show that IPNN prefers to put number 1,4,7,9 into one cluster and the rest into another cluster.
We also get a similar interesting results in CIPNN, as shown in~\figurename{~\ref{fig:1d_z_mnist}}, with stable number $\epsilon = 1$, the reconstructed image with 1-D latent space shows
a strong tendency to sort the categories into a certain order and the number 1,4,7,9 stays together in the latent space. Similar results are also found with 2-D latent space, see~\figurename{~\ref{fig:2d_z_mnist}}.  
Unfortunately, we currently do not know how to evaluate this sort tendency numerically.

\begin{figure}[htbp]
    \centerline{\includegraphics[width=1\linewidth]{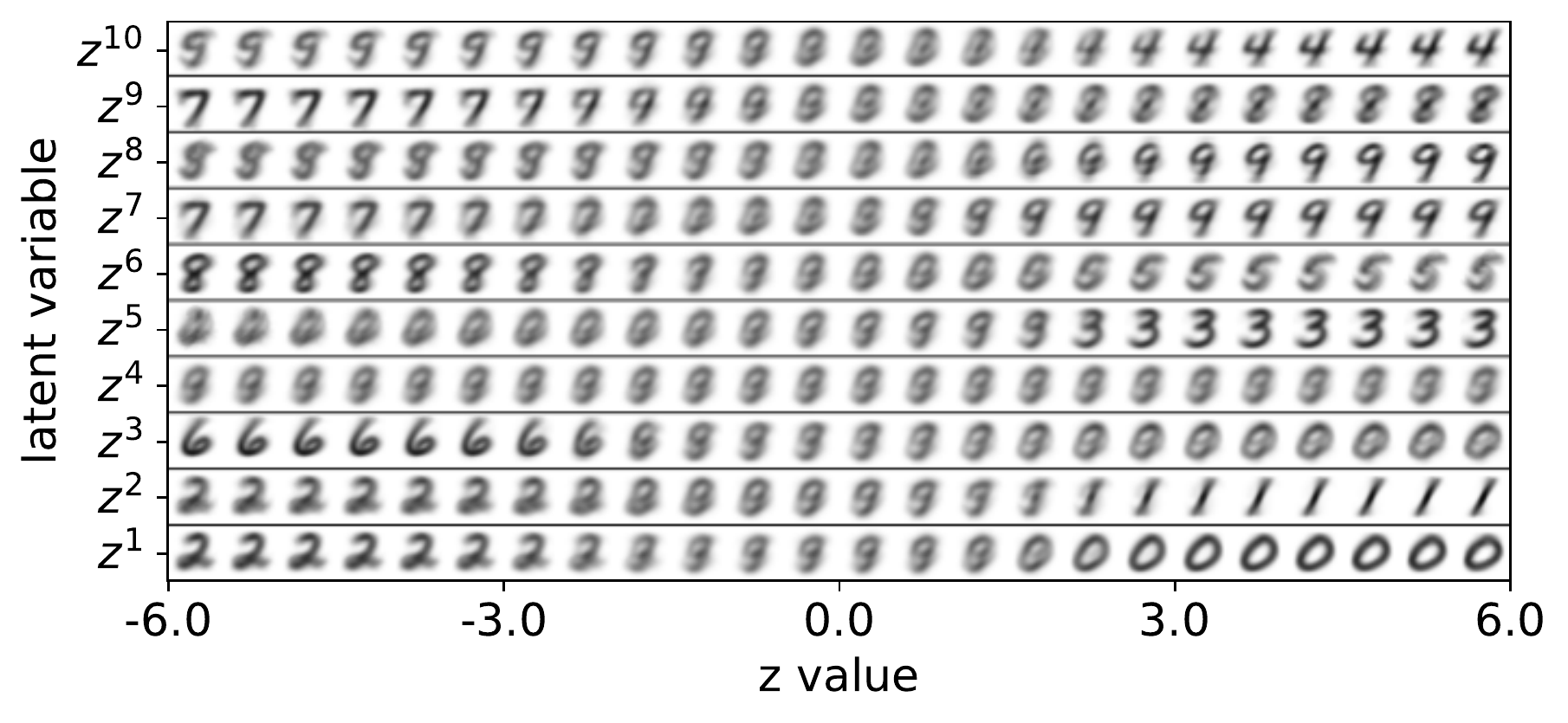}}
    \caption{Classification results of 10-D latent space: test accuracy is $97.1\%,\gamma=0.8, \epsilon\approx 0$. 
    Images are reconstructed with one latent variable $z^{i},i=1,2,\dots,10$, see Eq.~\eqref{eq:reconstructed_latent_image}.}
    \label{fig:10d_z_mnist}
    \end{figure}

With the visualization method proposed in Eq.~\eqref{eq:reconstructed_latent_image}, 
we can see what each latent variable has learned in the 10-D latent space.
As shown in~\figurename{~\ref{fig:10d_z_mnist}}, each latent variable focuses on mainly one or two different categories.

As shown in~\figurename{~\ref{fig:2d_z_mnist}}(a,d,g), with a proper regularization factor $\gamma$, the test dataset is mapped to a
relative small latent space, and the over-fitting problem is avoided.
Besides, in~\figurename{~\ref{fig:2d_z_mnist}}(b,e,h) each joint sample area correspond to one unique category, this is consistent with our Proposition~\ref{pps:convergence}.
In~\figurename{~\ref{fig:2d_z_mnist}}(c,f,i), the reconstructed image follows the conditional joint distribution $P \left (y_{l} \mid z^{1},z^{2} \right ), l = 0,2,\dots,9$.

\subsection{Results of Auto-Encoder Tasks}\label{sec:autoencoder_results}

\begin{figure}[htbp]
    \centering
        \begin{subfigure}[b]{0.49\linewidth}
                \centering
                \includegraphics[width=1\linewidth]{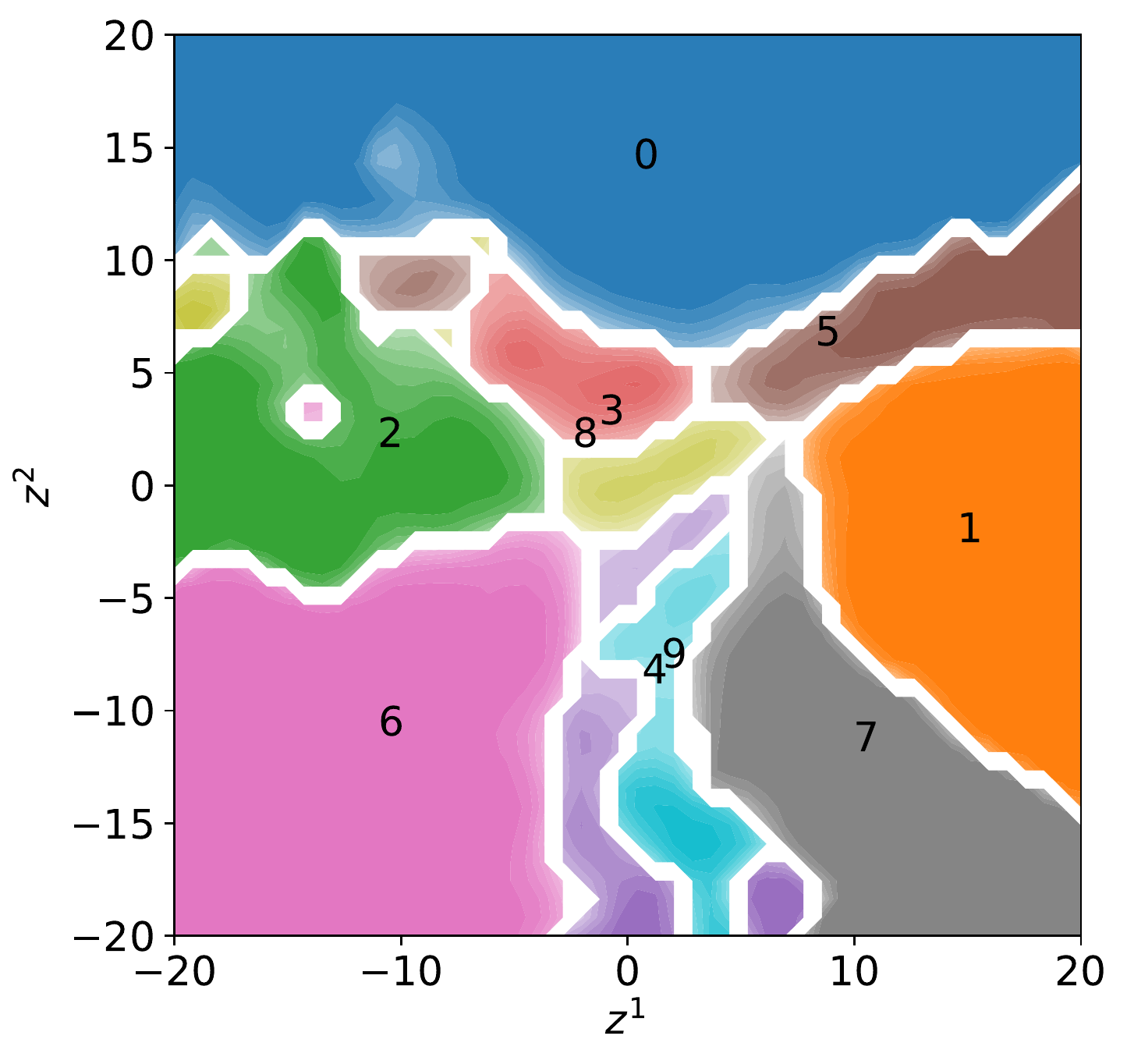}
                \caption{CIPAE model}
                \label{fig:2d_z_ae_mnist}
        \end{subfigure}
        \begin{subfigure}[b]{0.49\linewidth}
                \centering
                \includegraphics[width=1\linewidth]{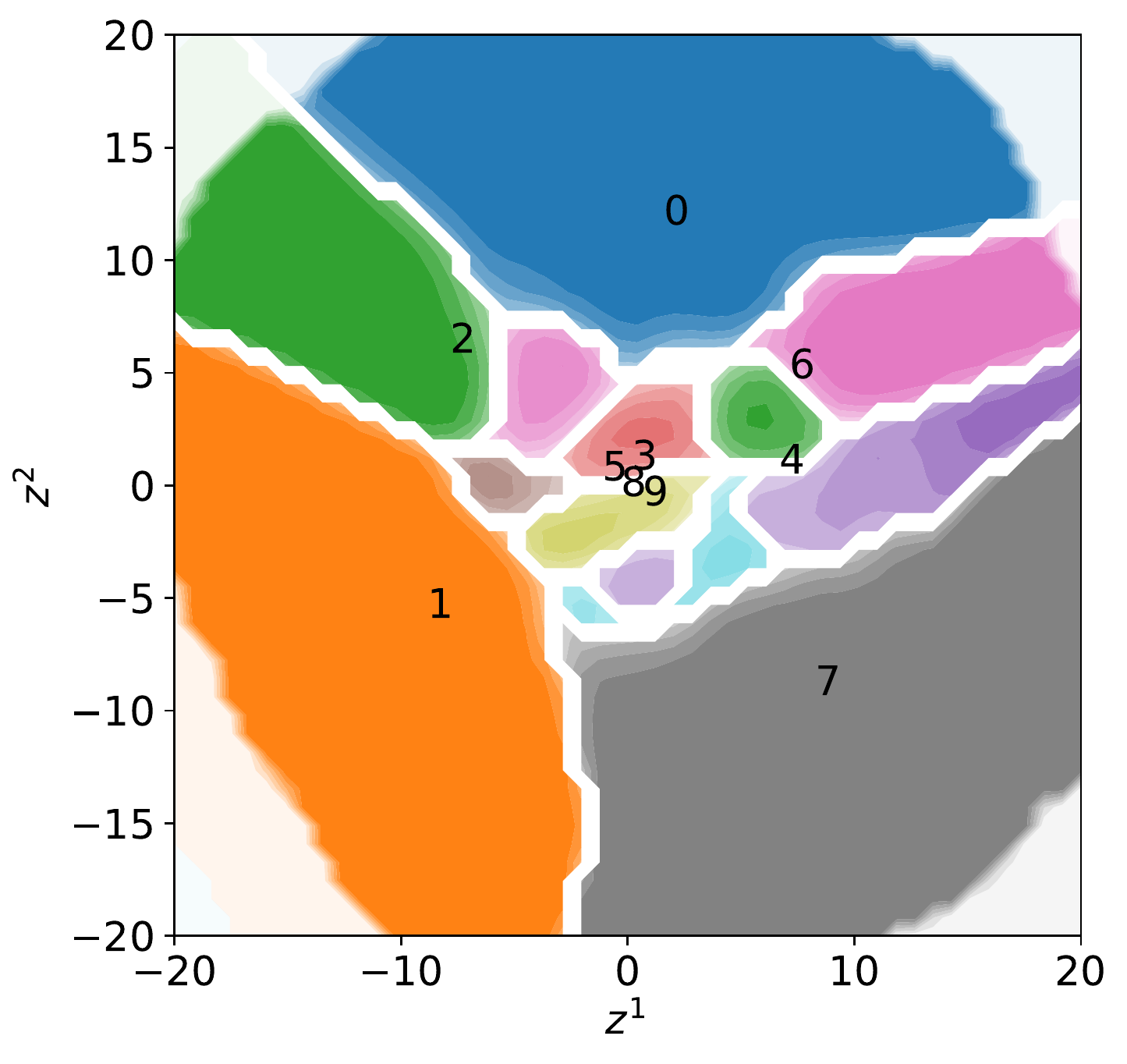}
                \caption{VAE model}
                \label{fig:2d_z_vae_mnist}
        \end{subfigure}
        \caption{Auto-encoder results of 2-D latent space evaluated with CIPNN model on MNIST\@: test accuracy is $70.1\%$ for CIPAE and $67.4\%$ for VAE, $\gamma=0.98, \epsilon\approx 0$\@.
        Conditional probability distribution of each category $P \left (y_{l} \mid z^{1},z^{2} \right ), l = 0,2,\dots,9$. Colors represent probability value: from 1-dark to 0-light;}
        \label{fig:2d_z_aevsvae_mnist}
    \end{figure}

In this section, we will make a comparison with our CIPAE and VAE~\cite{cipnn:VAE} model using train setup in~\figurename{~\ref{fig:train_process_b}},
for VAE model, we replace CIPAE part with VAE, in order to be able evaluate it with CIPNN model.
Besides, the regularization loss of VAE is switched to our proposed loss, see Eq.~\eqref{eq:regular_loss}.
As shown in~\figurename{~\ref{fig:2d_z_aevsvae_mnist}}, the results of auto-encoder tasks between CIPAE and VAE
are similar, this result further verifies that CIPAE is the analytical solution.

\section{Conclusion}\label{sec:conclusion}

General neural networks, such as FCN, Resnet~\cite{cipnn:resnet}, Transformer~\cite{cipnn:Transformer}, can be understood as a complex
mapping function $f:X\rightarrow Y$~\cite{cipnn:deep_learning}, but they are black-box for human~\cite{cipnn:blackbox_NN}.
Our proposed model can be understood as two part: $f:X\rightarrow Z$ and $P(Y \mid Z):Z\rightarrow Y$, the first part is still black box for us,
but the latter part is not unknown anymore. Such kind of framework may have two advantages: the first part can be used to detect the attributes of datasets and
summarize the common part of different categories, as shown in \figurename{~\ref{fig:1d_z_mnist}}; the latter part is a probabilistic model, which may be used to build a large
Bayesian network for complex reasoning tasks. 

Besides, our proposed framework is quite flexible, e.g. from $X$ to $Z$, we can use multiple neural networks with different structures to extract specific attributes as different random variables $z^{i}$,
and these random variables will be combined in the statistical phase.

Although our proposed model is derived from indeterminate probability theory, we can see Determinate from the expectation form in Eq.~\eqref{eq:P_Y_X_via_Z_1}.
Finally, we'd like to finish our paper with one sentence: The world is determined with all Indeterminate.

\section*{Acknowledgment}

I'd like to thank Chuang Liu, Xiaofeng Ma, Weijia Lu, Ning Wu, Bingyang Li,
Zhifei Yang, Peng Liu, Lin Sun, Xiaodong Zhang and Can Zhang for their review of this paper.

\appendix

\section{Visualization}\label{app:gamma}

\begin{figure}[htbp]
    \centerline{\includegraphics[width=1\linewidth]{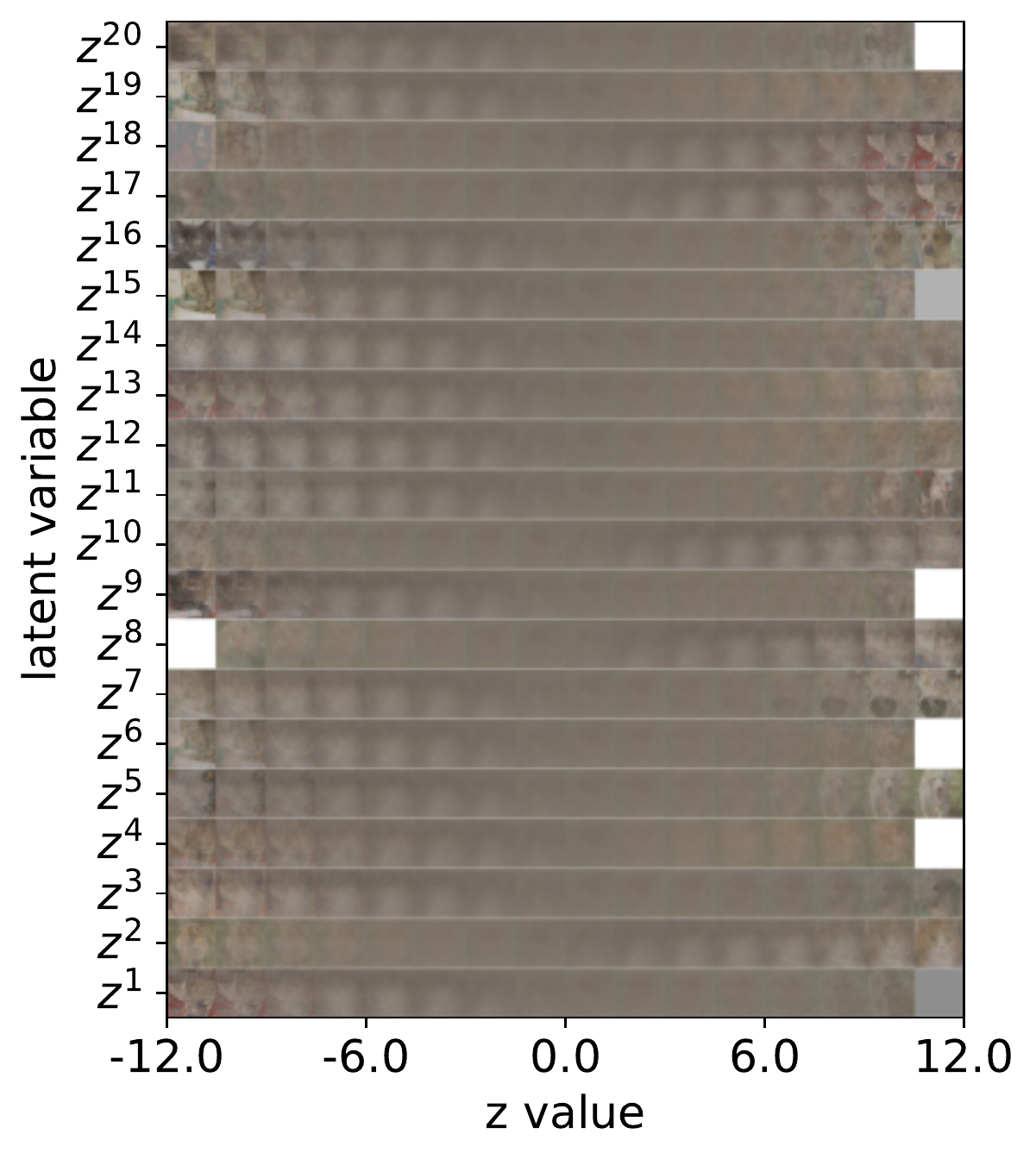}}
    \caption{Classification results of 20-D latent space on Dogs-vs.-Cats-Redux: test accuracy is $95.2\%,\gamma=0.999, \epsilon\approx 0$. 
    Images are reconstructed with one latent variable $z^{i},i=1,2,\dots,20$, see Eq.~\eqref{eq:reconstructed_latent_image}.}
    \label{fig:20d_z_petimages}
    \end{figure}

\figurename{~\ref{fig:20d_z_petimages}} shows classification results of 20-D latent space on Dogs-vs.-Cats-Redux dataset,
we can see that each latent variable focuses on both two different categories.

\begin{figure}[htbp]
    \centering
        \begin{subfigure}[c]{0.47\linewidth}
                \centering
                \includegraphics[width=1\linewidth]{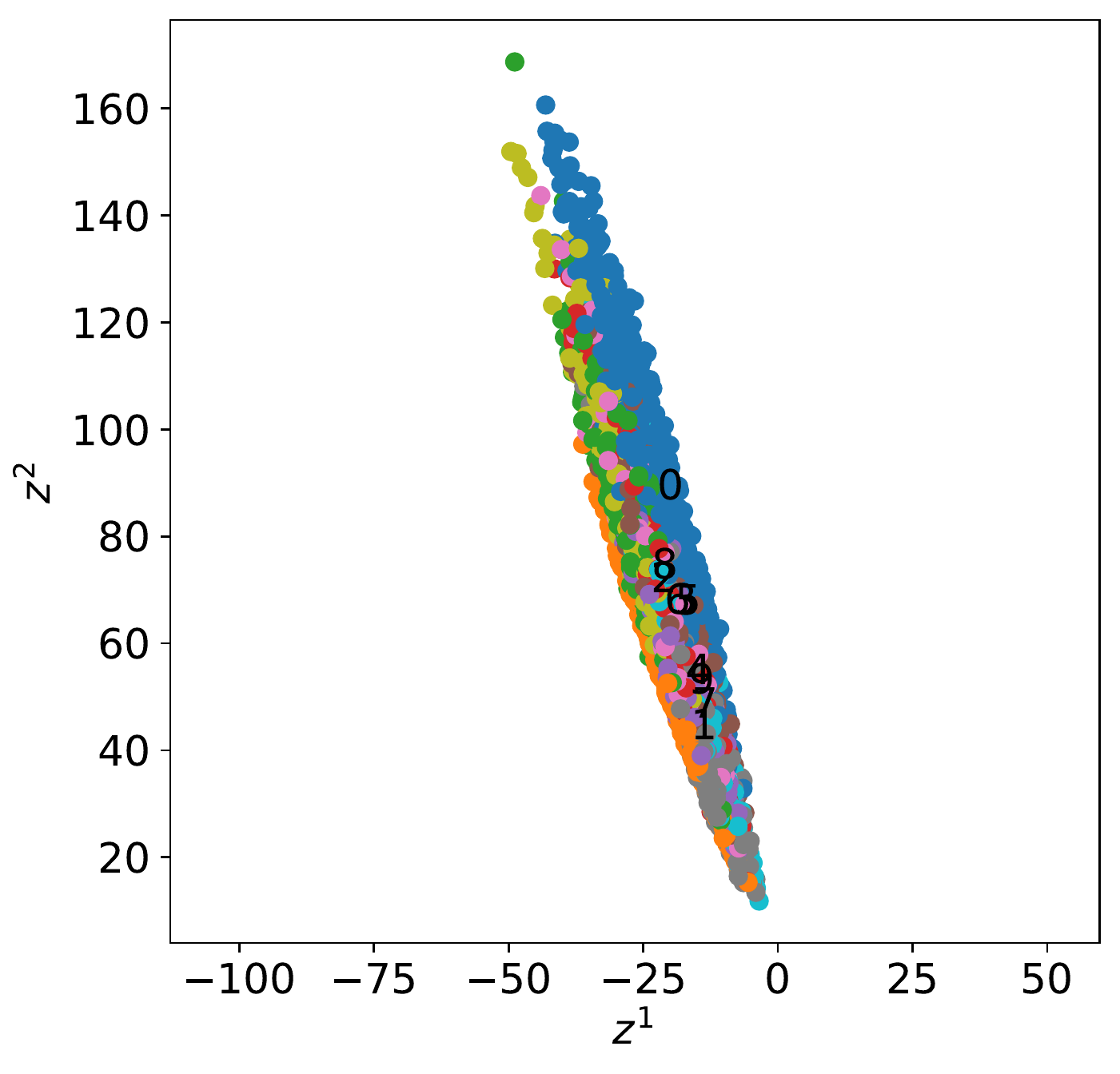}
                \caption{w/o. $\mathcal{L}_{2}, acc = 9.8\%$}
        \end{subfigure}
        \begin{subfigure}[c]{0.49\linewidth}
                \centering
                \includegraphics[width=1\linewidth]{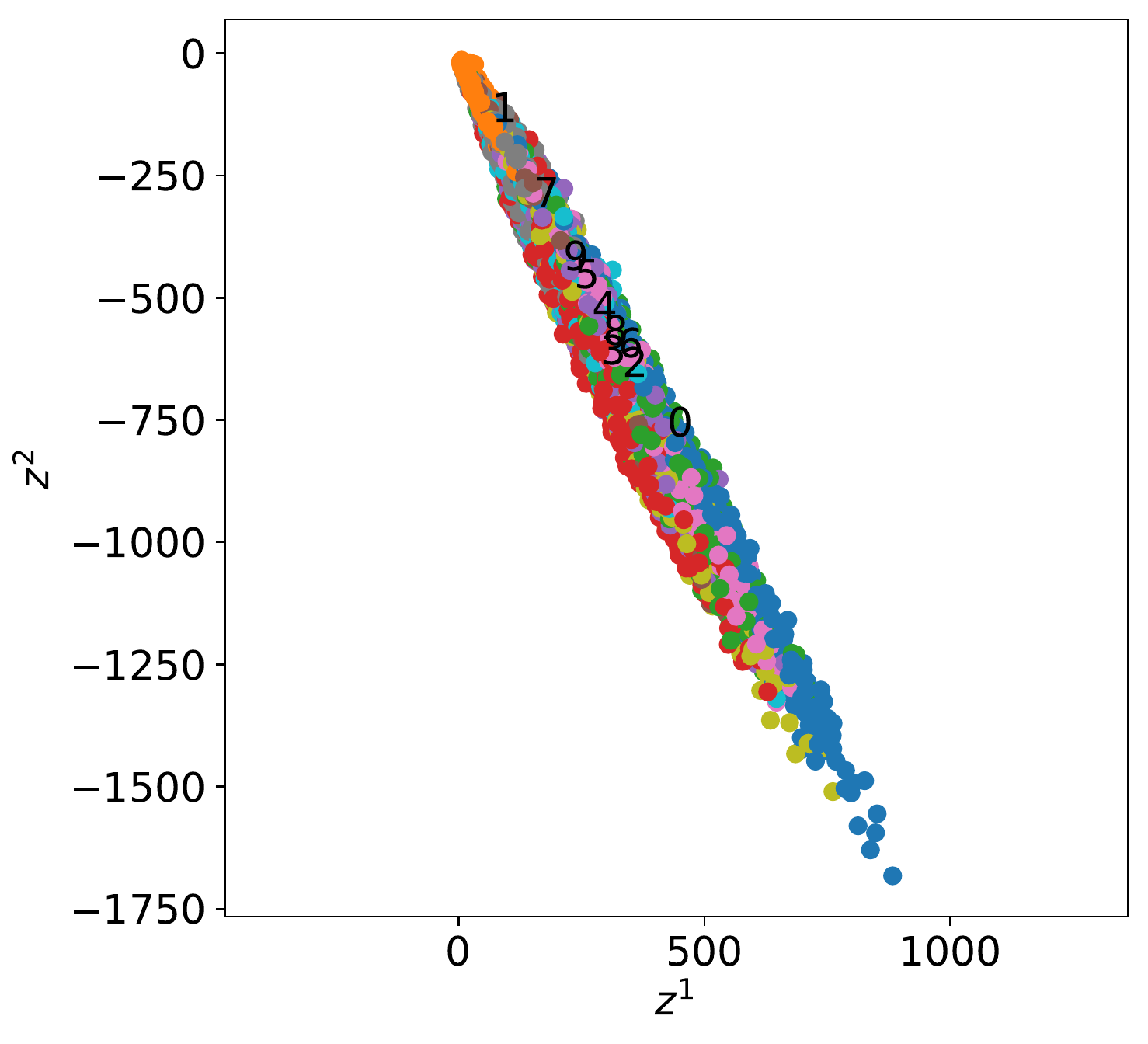}
                \caption{$\gamma = 1, acc = 25.1\%$}
        \end{subfigure}

        \begin{subfigure}[c]{0.49\linewidth}
            \centering
            \includegraphics[width=1\linewidth]{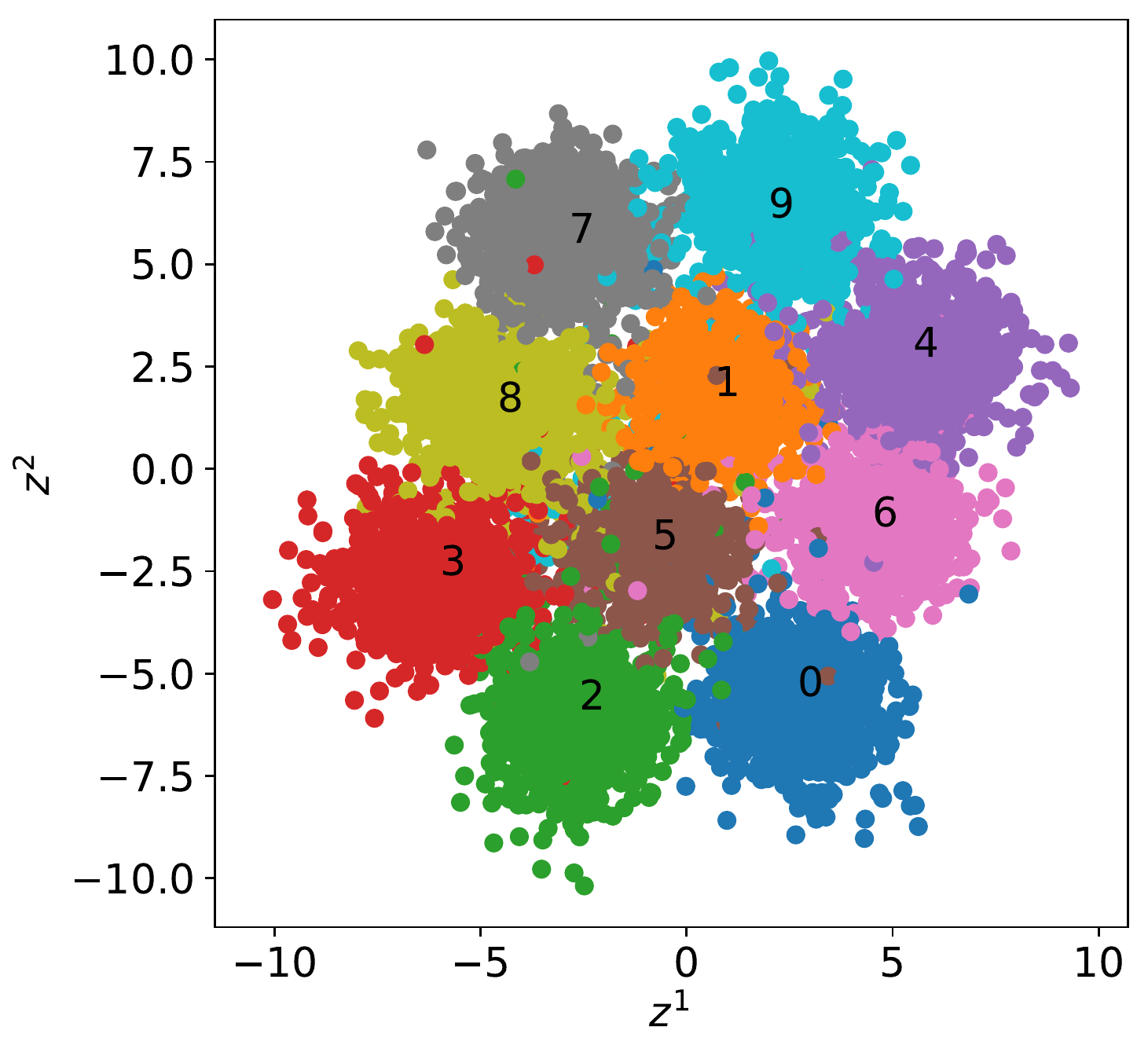}
            \caption{$\gamma = 0.9, acc = 96.2\%$}
        \end{subfigure}
        \begin{subfigure}[c]{0.46\linewidth}
            \centering
            \includegraphics[width=1\linewidth]{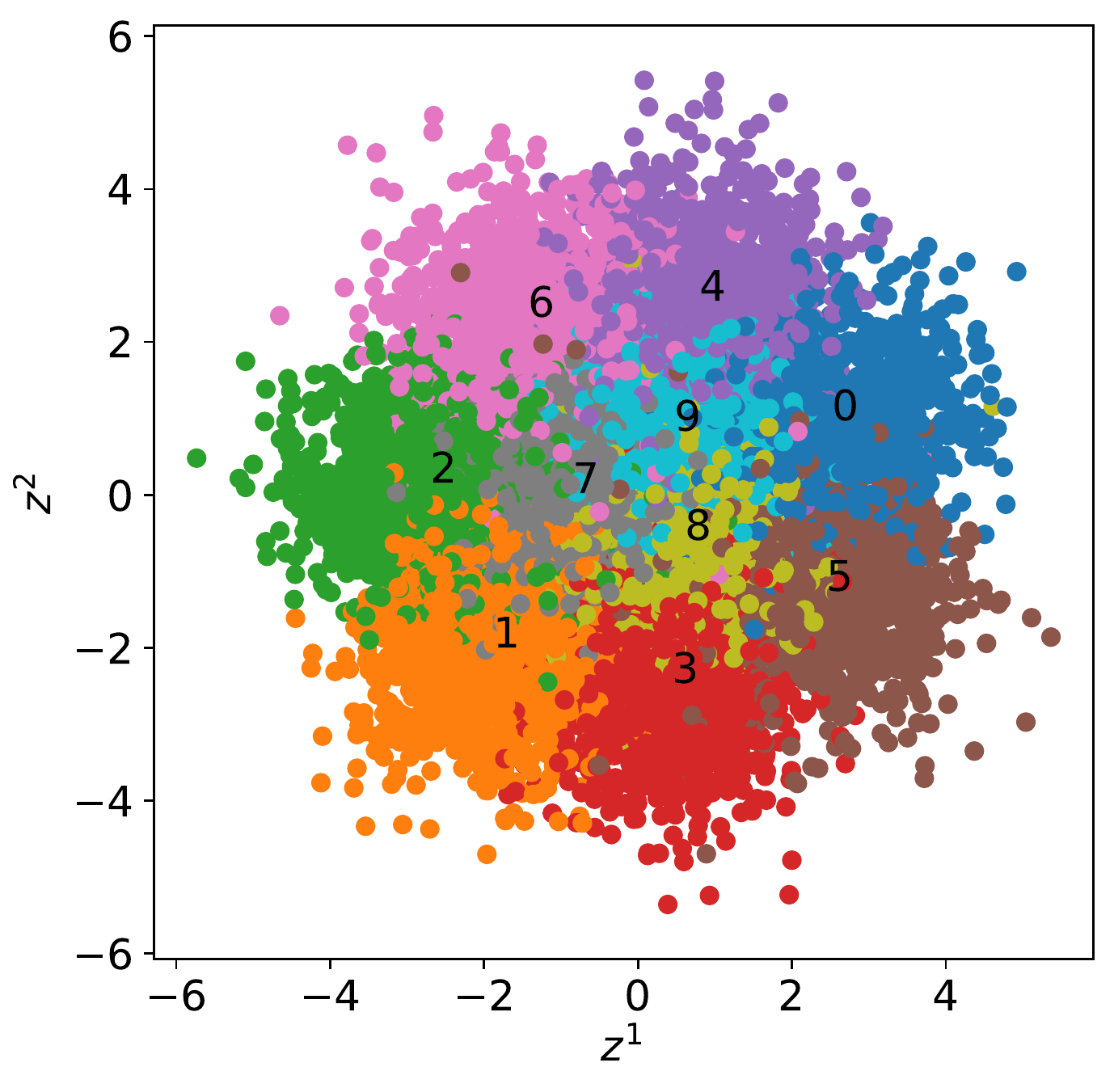}
            \caption{$\gamma = 0.6, acc = 84.6\%$}
        \end{subfigure}

        \begin{subfigure}[c]{0.49\linewidth}
                \centering
                \includegraphics[width=1\linewidth]{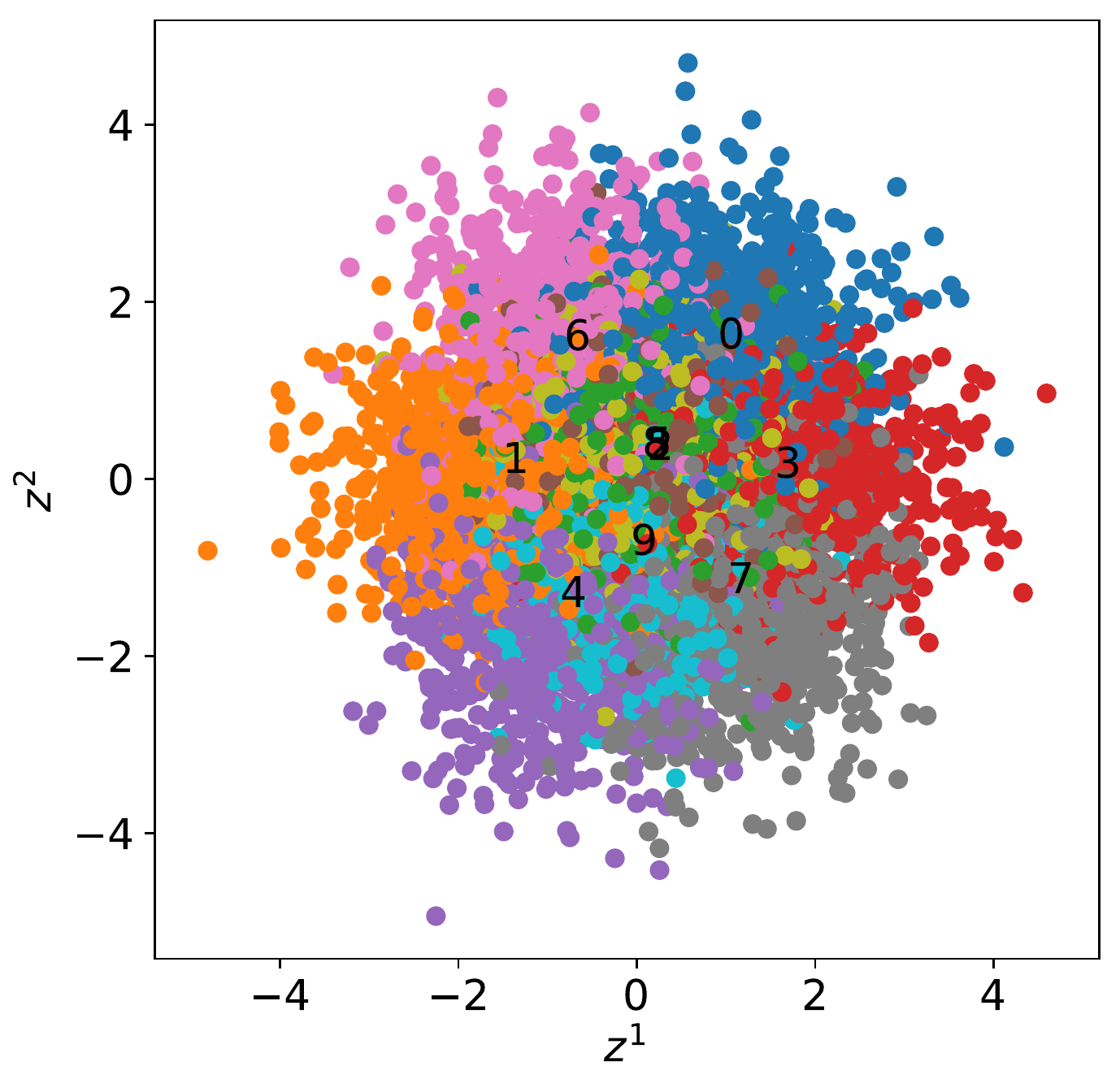}
                \caption{$\gamma = 0.3, acc = 54.8\%$}
        \end{subfigure}
        \begin{subfigure}[c]{0.49\linewidth}
            \centering
            \includegraphics[width=1\linewidth]{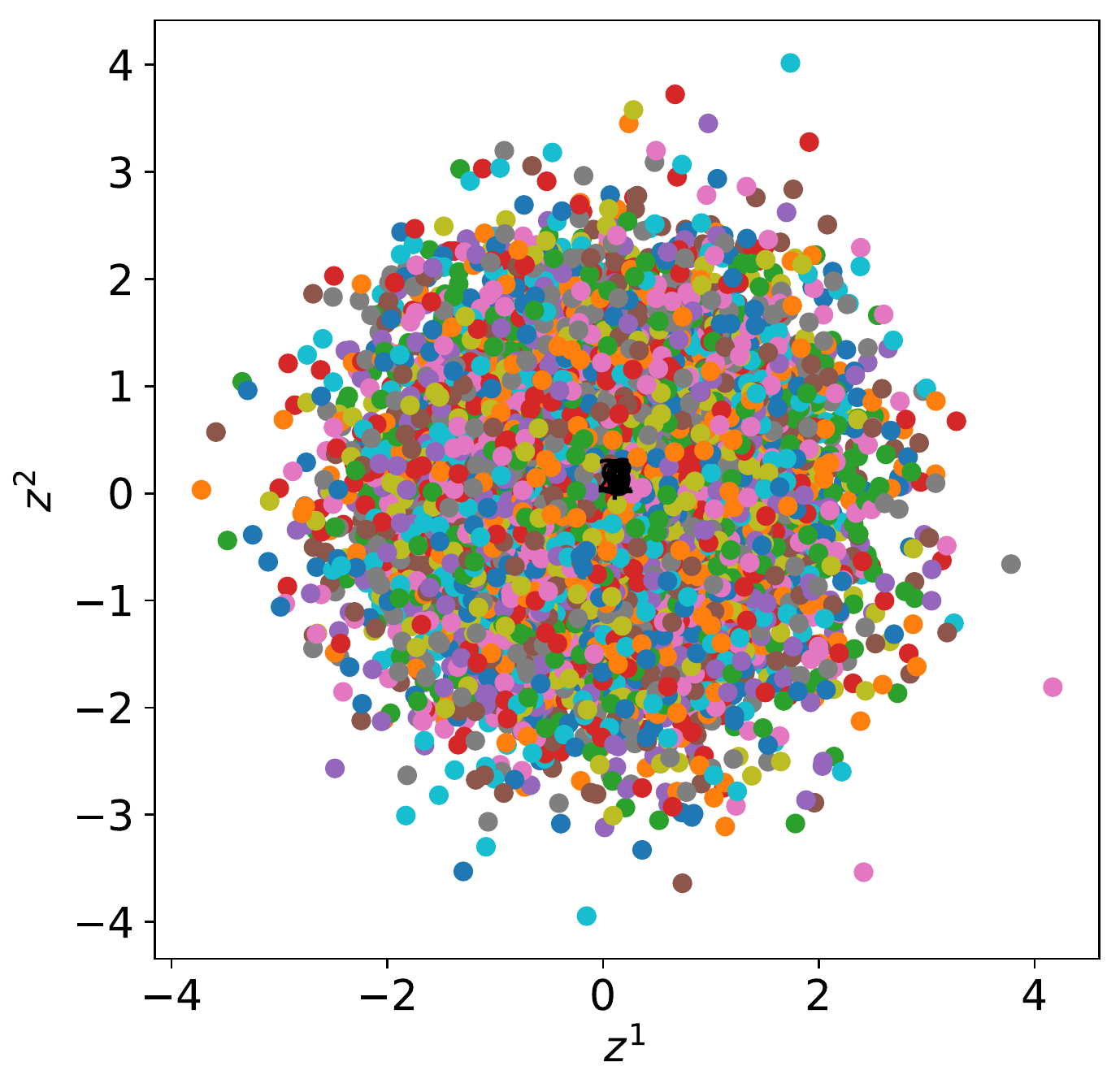}
            \caption{$\gamma = 0, acc = 11.4\%$}
        \end{subfigure}
        \caption{Impact analysis of regularization factor $\gamma$ with 2-D latent space of classification results on MNIST\@.
                (a) Without regularization term, the model has a strong over-fitting problem, and the variance is near to 0,
                    that's why it's space is not big than (b).
                (b) The variance is around 1, however, without constrain the distribution to be connected together,
                    the space is very high, and it also shows over-fitting problem.
                (c) With proper $\gamma$ value, we get an optimal effective latent space.
                (d,e,f) As $\gamma$ value further reduces, the latent space is getting smaller, and both tain and test accuracy is reducing due to over regularization. }
        \label{fig:2d_z_mnist_gammas}
        \end{figure}

Impact analysis of regularization factor $\gamma$ is discussed in~\figurename{~\ref{fig:2d_z_mnist_gammas}}.

{\small
\bibliographystyle{ieee_fullname}
\bibliography{cipnn}
}

\end{document}